\documentclass[11pt]{article}

\usepackage[preprint]{acl}

\usepackage{times}
\usepackage{latexsym}
\usepackage{booktabs}

\usepackage[T1]{fontenc}

\usepackage[utf8]{inputenc}

\usepackage{microtype}

\usepackage{inconsolata}

\usepackage{algorithm}
\usepackage{amsmath}
\usepackage{algpseudocode}

\usepackage{graphicx}
\usepackage{multirow}

\usepackage{bm}

\usepackage[dvipsnames]{xcolor}

\usepackage[most]{tcolorbox}
\usepackage{xcolor}

\definecolor{origcolor}{HTML}{5F5E5A}
\definecolor{baselinecolor}{HTML}{A32D2D}
\definecolor{seamscolor}{HTML}{534AB7}

\newtcolorbox{origbox}[1][]{%
  colback=white,
  colframe=origcolor,
  boxrule=0.5pt,
  arc=2pt,
  left=6pt, right=6pt, top=4pt, bottom=4pt,
  fonttitle=\small\bfseries,
  coltitle=origcolor,
  attach boxed title to top left={xshift=8pt, yshift=-7pt},
  boxed title style={colback=white, colframe=white, boxrule=0pt},
  title={Original},
  #1
}

\newtcolorbox{baselinebox}[1][]{%
  colback=white,
  colframe=baselinecolor,
  boxrule=0.5pt,
  arc=2pt,
  left=6pt, right=6pt, top=4pt, bottom=4pt,
  fonttitle=\small\bfseries,
  coltitle=baselinecolor,
  attach boxed title to top left={xshift=8pt, yshift=-7pt},
  boxed title style={colback=white, colframe=white, boxrule=0pt},
  title={Cluster-Summary-Centric},
  #1
}

\newtcolorbox{seamsbox}[1][]{%
  colback=white,
  colframe=seamscolor,
  boxrule=0.5pt,
  arc=2pt,
  left=6pt, right=6pt, top=4pt, bottom=4pt,
  fonttitle=\small\bfseries,
  coltitle=seamscolor,
  attach boxed title to top left={xshift=8pt, yshift=-7pt},
  boxed title style={colback=white, colframe=white, boxrule=0pt},
  title={SEAMS},
  #1
}

\title{Reducing Authorship Attribution Risk in Student Texts Through Style-Conditioned Rewriting}

\author{
 \textbf{Li Siyan}\thanks{Co-first authors},
 \textbf{Zhen Xu}\footnotemark[1],
 \textbf{Jiayi Chen},
 \textbf{Shopnil Shahriar},
 \textbf{Alessandro Sayad},
 \textbf{Xuanming Zhang}\\
 \textbf{Renzhe Yu}\thanks{Co-senior authors},
  \textbf{Zhou Yu}\footnotemark[2]
\\
 Columbia University
\\
 \texttt{\href{mailto:siyan.li@columbia.edu}{siyan.li@columbia.edu},\href{mailto:zx2393@tc.columbia.edu}{zx2393@tc.columbia.edu}
 }
}

\begin{document}
\maketitle
\begin{abstract}
Building NLP applications for education is constrained by data scarcity due to student privacy concerns and data regulations. To address this, we propose a style-aware approach to rewriting student forum posts that retains analytical value while reducing identifiability. Our method isolates content, structure, and style as separate conditioning channels and aggregates over student-disjoint exemplar partitions within style clusters, extending Differentially Private In-context Learning (DP-ICL) with cluster-restricted retrieval. Across three educational NLP datasets, our approach achieves high style fidelity, preserves downstream task performance, and reduces authorship attribution risk relative to the original student-written texts, including against a dedicated authorship-attribution model.

\end{abstract}

\section{Introduction}

NLP applications for education face significant \textbf{reproducibility challenges}: real student data is subject to strict data protection requirements under privacy regulations and institutional policies \cite{pardo2014ethical,young2014educational,neshaei2025bridging,lohse2019surveying}. We note that simply redacting sensitive information is insufficient, as Institutional Review Boards (IRBs) often prohibit the release of student-authored content; therefore, even student texts without any personally-identifiable information (PII) cannot be released. Given these constraints, researchers have investigated general-purpose LLMs for student simulation \cite{jin2025teachtune,li2026type,xu2025classroom} and data synthesis \cite{li2025balancing,kieser2023educational}.

A central challenge for current simulation and synthesis efforts is the \textit{tradeoff between privacy and utility}: methods that strongly protect identity tend to degrade the linguistic and pedagogical signals that make the data useful. Our paper addresses this tradeoff in the context of \textbf{discussion forum posts}.

\begin{figure}[!t]
    \centering
    \includegraphics[width=\linewidth]{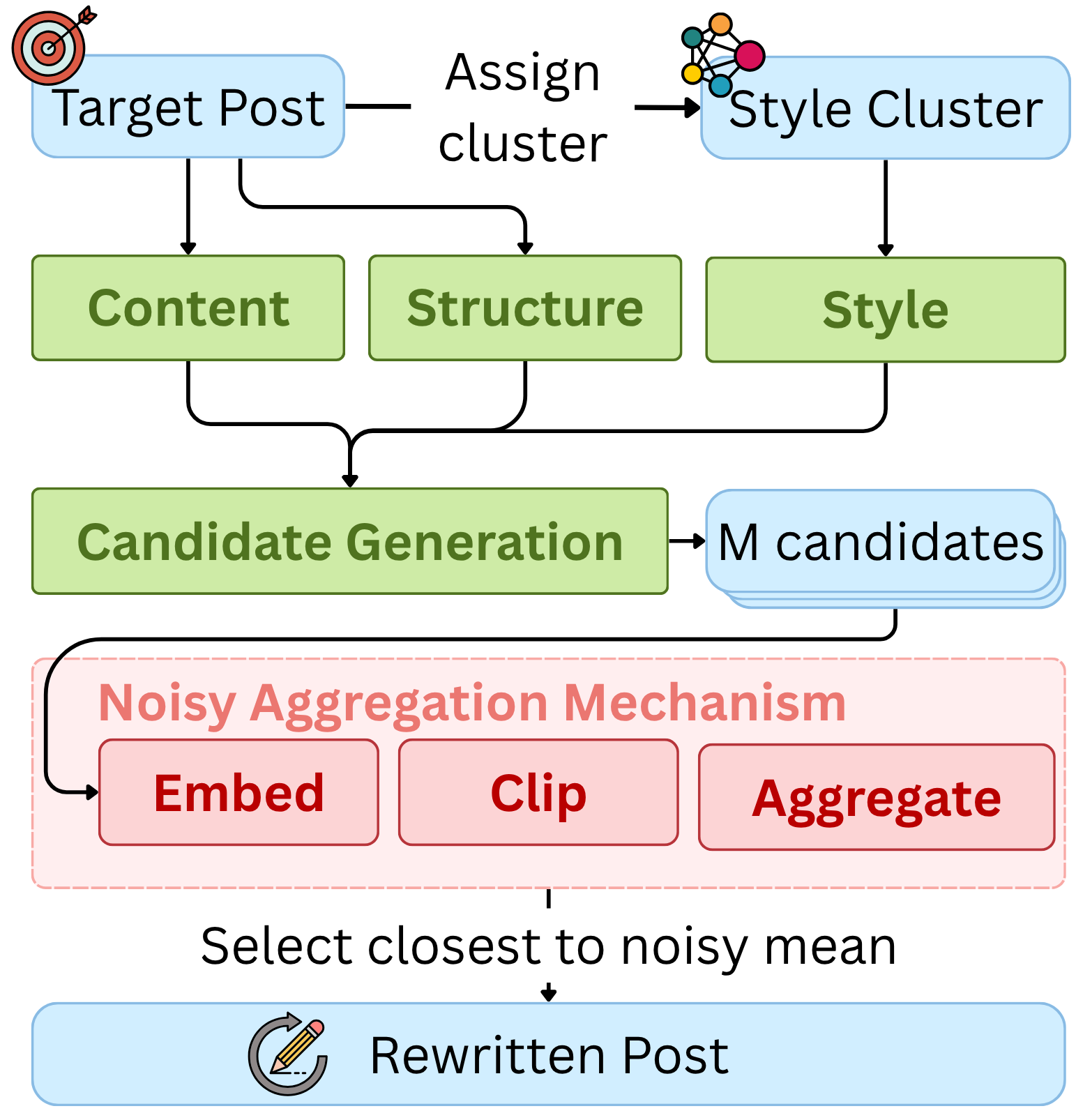}
    \caption{%
  \textbf{SEAMS pipeline.}
  Three conditioning channels (bullet summary, template, cluster exemplars) condition $M$ parallel LLM generations.
  Candidates are embedded, $\ell_2$-clipped, and averaged with Gaussian noise; the released rewrite is the candidate closest to the mean.
}
\label{fig:pipeline}
\end{figure}

Discussion forums are the de facto interactive medium for both MOOCs and university courses, where learners ask questions, discuss ideas, and reflect on course material \cite{onyema2019online}. Unlike formal academic writing, forum posts embody naturalistic student language, reflecting diverse cognitive and communicative tendencies as evidenced by their varying lengths, tones, and rhetorical structures. These posts are rich with pedagogically meaningful signals, including urgency, confusion, and emotional states \cite{khodeir2021bi,agrawal2015youedu}, making them a valuable, rich resource
\cite{cho2024systematic}. More recently, increasingly capable language models have spurred the deployment of virtual teaching assistants in discussion forums \cite{goel2018jill,shah2024students,kweon2025large}, 
which require training and evaluation data that reflects the full distribution of student writing styles.

In the context of student simulation, general-purpose models often fail to capture the diversity in personalities, cognitive capabilities, and stylistic idiosyncrasies \cite{martynova2025can,zheng2025teaching}. This failure could lead to high-stakes consequences; a virtual teaching assistant trained on distributionally flat synthetic student data may fail to produce the pedagogically optimal response in deployment.
Addressing this requires synthetic data that captures the underlying stylistic distribution of real student populations.

We propose \textit{rewriting} each real student post into a less-identifiable version that preserves the stylistic and semantic properties needed for downstream NLP; unlike fully synthetic generation, each output corresponds to exactly one source post. Prior work on synthetic data \cite{li2023synthetic,naduas2025synthetic} and differentially private language modeling \cite{yu2021differentially,shi2022just,flemings2024differentially} pays little attention to style-preserving synthesis in education. The core challenge is to \textbf{preserve cluster-level style for utility while suppressing individual stylistic fingerprints} that enable re-identification. We address this by decomposing student texts into content, structure, and style, then aggregating over student-disjoint exemplar partitions to limit any one author's impact. We implement this via Style-conditioned Ensemble Aggregation with Multi-channel Synthesis (\textbf{SEAMS}). SEAMS extends DP-ICL (\citealp{wu2024privacy}, more details in Section \ref{sec:pmg}) by incorporating stylistically similar exemplars while restricting influence of individual authors. Across three datasets, SEAMS preserves style, maintains downstream task performance, and empirically reduces authorship attribution risk relative to the originals.

Our contributions are as follows: (1) \textbf{SEAMS}\footnote{Code, including all prompts, and the SEAMS-rewritten Stanford MOOC and Stack Exchange data are released here: \url{https://github.com/siyan-sylvia-li/seams_release}; the proprietary dataset cannot be released under its data agreement.}, a privacy-motivated, style-aware pipeline for rewriting student forum posts.
(2) Evidence across three educational NLP datasets that SEAMS preserves downstream task performance and high style fidelity.
(3) Evidence that SEAMS empirically reduces authorship attribution risk compared to original student posts.





\section{Related Work}
Recent work has increasingly turned to LLMs for simulating students, both for training teachers \cite{mikeska2025generative}, evaluating pedagogical agents \citep{jin2025teachtune}, modeling behavioral trajectories \citep{xu2025classroom}, and supporting authentic human-AI collaboration in educational settings \citep{li2026type}. However, these simulators rely on prompting general-purpose LLMs, which may not faithfully reproduce the heterogeneity of real student populations \cite{wu2025embracing,martynova2025can} and offer no mechanism for releasing privacy-protected versions of real student writing. SEAMS addresses both.

\noindent \textbf{Style-conditioned Generation:} Prior literature on conditional generation \cite{zhang2023survey} and style transfer has studied approaches to steer LLM output style, from exemplar-based prompting \cite{reif2022recipe}, contrastive prompting \cite{suzgun2022prompt}, and controlled text editing \cite{liu2024step,luo2023prompt}. SEAMS draws on this literature but targets a privacy-preserving rather than purely stylistic objective.


\noindent \textbf{Differentially Private Text Generation:}
Training-time DP methods extend DP-SGD \citep{abadi2016} to language model fine-tuning \citep{yu2021differentially, shi2022just} and attribute-conditioned synthetic text \citep{yue2023, mattern2022}, or train the generator itself from DP signals \citep{tan2025ctcl, hou2025popri, xu2026dprft}. Inference-time methods instead release each token under DP \citep{tang2024privacy, amin2024private, vinod2025invisibleink,flemings2024differentially}, refine samples through API calls with DP-compliant feedback \citep{xie2024augpe, chien2026maple, wang2025secpe}, or aggregate predictions across disjoint demonstration sets, as in DP-ICL \citep{wu2024privacy}. Generally, these methods produce a new \textit{dataset} with guarantees, while SEAMS studies one-to-one rewriting, adding multi-channel conditioning and cluster-restricted retrieval, and trade formal guarantees for style fidelity. The closest one-to-one method, DP-Prompt \citep{utpala2023locally}, paraphrases each document under local DP.

\section{SEAMS}

SEAMS isolates three conditioning channels from student texts: content, structure, and style. It then aggregates over candidate generations from student-disjoint exemplar partitions within style clusters. The rest of this section describes each component. \textbf{SEAMS does not include PII redaction}; therefore, any institution interested in using SEAMS should perform redactions independently.



\subsection{Preprocessing}

Each post is converted into two SEAMS input channels: bullet summaries for content and academic-content-removed templates for structure. Together with retrieved cluster exemplars for style, these form the three conditioning channels.

\noindent \textbf{Bullet summaries:} We summarize each post with \texttt{Qwen-2.5-7B-Instruct} \cite{qwen2024qwen2} into bullet points that retain content while removing style signals, using a GEPA-optimized \cite{agrawal2025gepa} prompt. We use this model for cost considerations, and a 7B model is sufficient for this task. The optimization uses Stack Exchange posts, maximizing semantic similarity, stylistic \textit{dissimilarity}, reconstruction fidelity, and succinctness. Details are in Appendix \ref{app:bullet_summarizer}.

\noindent \textbf{Academic-content-removed templates:} We also use GPT-5-mini (\textbf{permitted by our IRB}) to convert each post into a template where academic content is replaced by bracketed slots, e.g., \texttt{[concept or knowledge]}, while preserving connectives and rhetorical structure. We chose GPT-5-mini because templating requires reliable span-level identification of domain-specific content; in pilot comparisons, GPT-5-mini produced templates that matched expert annotations more closely than \texttt{Qwen3-30B-A3B-Instruct-2507}. The template was originally designed by education collaborators for clustering, but we find it suitable for SEAMS' structural channel. We GEPA-optimize the templating prompt against expert annotations on proprietary student posts, maximizing the per-character agreement rate between model predictions and human reference (Appendix~\ref{app:rr_qa}).




\subsection{Embedding Model Selection}
Well-formed, meaningful style clustering requires an embedding model that distinguishes posts by writing style rather than topic. We compare \texttt{StyleDistance} \cite{patel2025styledistance}, \citet{wegmann2022same}, Coh-Metrix \cite{graesser2004coh}, and \texttt{all-Mini-LM-v6} \cite{reimers2019sentence}. \texttt{StyleDistance} is selected because it separates within-student and between-student post pairs the most in terms of embedding cosine similarities, yielding the highest t-test statistics (Appendix \ref{app:embedding_selection}).

\subsection{Per-Seed Style Clustering}

HDBSCAN (with UMAP reduction first) is selected for its ability to identify clusters of varying density without requiring a fixed cluster count. 
We sweep a combination of HDBSCAN settings, with minimum cluster size $mcs \in \{100, 150, 200\}$ and minimum samples $ms \in \{10, 15, 20, 30\}$. The best HDBSCAN setting for the dataset is selected and fixed across seeds based on the average validation cluster membership similarity. Intuitively, high cluster membership similarity indicates better similarity between the unseen posts and existing clusters, leading to higher-quality retrieval.

\subsection{Cross-Seed Meta-Cluster Formation}
Per-seed HDBSCAN clusters can be unstable across random seeds. To identify a shared meta-cluster mapping across seeds for easy comparison (and to enable comparison against other cluster-based approaches), we perform average-linkage agglomerative hierarchical clustering to connect clusters from the previous step to form meta-clusters. As all data points are discussion forum posts, they occupy a similar style embedding space, potentially flattening the cosine similarity between cluster centers and reducing modularity in the final meta-cluster hierarchy. Therefore, we remove dominant shared directions using All-but-the-Top (ABTT, \citet{mu2017all}) on the embeddings. The final meta-cluster count aims to maximize cross-seed cluster representation, such that most meta-clusters aggregate clusters from multiple seeds rather than a single seed, enhancing generalizability. 

\subsection{Privacy-Motivated Generation}
\label{sec:pmg}

Once the clusters are finalized, the generation step consists of three components: student-disjoint partition, candidate generation, and noisy aggregation. We adopt the DP-ICL framework \cite{wu2024privacy} as the structural basis for this step. DP-ICL obtains differentially private responses from a noisy consensus over LLM outputs conditioned on disjoint exemplar subsets: noisy voting over predicted labels for classification, and noisy aggregation in an embedding or keyword space for generation. SEAMS follows the generation setting and aggregates in embedding space. We extend it with cluster-restricted retrieval and student-disjoint partitioning to match our specific threat model. Unlike DP-ICL's Embedding Space Aggregation (ESA), which returns a \textbf{public} zero-shot candidate \citep{wu2024privacy}, we select among the privately conditioned candidates themselves and retrieve exemplars by style similarity rather than random subsampling. We therefore \textbf{do not claim any formal differential privacy guarantees}; our framework is privacy-motivated, drawing structural principles from DP-ICL. We include ESA as a baseline. Algorithm~\ref{alg:seams} describes this procedure.

\noindent \textbf{Student-disjoint Partitioning:} Retrieved exemplars are divided into disjoint sets, and if a student authors multiple posts among the exemplars, they are all placed in the same partition. This is to ensure that any single student's individual style influences appear in only one of the $\mathcal{M}$ candidates, as one candidate is produced per partition. We use cluster-restricted retrieval (retrieving within the same cluster only), which further pulls candidates toward the \textit{cluster}'s typical register.

\noindent \textbf{Candidate Generation:} Given the partitions, one candidate rewrite is created by an LLM per partition; the content and structural channels are shared across candidates so that candidate differences stem from stylistic conditioning alone. The rewriter prompt is selected from alternatives proposed by the MIPROv2 prompt optimizer \cite{opsahl2024optimizing} under a simplified setting rather than refinement (selection details in Appendix~\ref{app:mipro}; the final prompt is given in Appendix~\ref{app:dspy_prompt}), as iterative refinement may steer \textit{all} generations toward the typical structure of forum posts to maximize average metric value, leading to low-diversity generations. 

\noindent \textbf{Noisy Aggregation:} Following DP-ICL, candidate embeddings from \texttt{StyleDistance} \cite{patel2025styledistance} are \textbf{$\ell_2$-clipped} to bound the influence of any single candidate, \textbf{averaged} to find a consensus, and \textbf{perturbed} with Gaussian noise to reduce the determinism in the selection. The chosen rewrite candidate is the one whose clipped embedding lies closest to the mean. Together, these operations limit the per-candidate influence on the output. 

Component ablations (Appendix~\ref{sec:ablations}) show that selecting the candidate most similar to mean embeddings filters low-quality tail outputs, while cluster restriction and noisy aggregation preserve fidelity and serve structural roles. 

\begin{algorithm}
\caption{Privacy-Motivated Generation}
\label{alg:seams}
\begin{algorithmic}[1]
\Require Target post $p$, meta-cluster $\mathcal{C}$, content bullets $b$,
         structural template $t$, retrieval size $K$, partition count $M$,
         noise multiplier $\sigma$, clipping bound $\tau$
\Ensure Synthetic post $\hat{p}$
\State $\mathcal{E} \leftarrow \textsc{TopK}(p, \mathcal{C}, K)$
\Comment{$K$ cluster members ranked by cosine similarity to $p$}
\Statex{$\triangleright$ \textbf{\textit{Student-Disjoint Partitioning}}}
\State $\{\mathcal{E}_1, \dots, \mathcal{E}_M\} \leftarrow \textsc{DisjPartition}(\mathcal{E}, M)$

\For{$m = 1$ \textbf{to} $M$}
    \Statex{\ \ \ \ \ \ $\triangleright$ \textbf{\textit{Candidate Generation}}}
    \State $c_m \leftarrow \textsc{LLM}\left(\textsc{Assemble}(b, t, \mathcal{E}_m)\right)$
    \State $\mathbf{e}_m \leftarrow \textsc{Embed}(c_m)$
    \State $\hat{\mathbf{e}}_m \leftarrow \mathbf{e}_m \cdot \min\!\left(1, \frac{\tau}{\|\mathbf{e}_m\|_2}\right)$
    \Comment{clip to $\tau$}
\EndFor
\Statex{$\triangleright$ \textbf{\textit{Noisy Aggregation}}}
\State $\bar{\mathbf{a}} \leftarrow \frac{1}{M} \sum_{m=1}^{M} \hat{\mathbf{e}}_m + \mathcal{N}\!\left(\mathbf{0}, \left(\tfrac{\sigma \tau}{M}\right)^2 \mathbf{I}\right)$

\State $m^* \leftarrow \arg\min_m \|\hat{\mathbf{e}}_m - \bar{\mathbf{a}}\|_2$
\State \Return $\hat{p} \leftarrow c_{m^*}$
\end{algorithmic}
\end{algorithm}


\section{Datasets}
We validate SEAMS across three complementary, English-language datasets spanning different educational contexts and downstream classification tasks (Table~\ref{tab:dataset_stats}). 

\subsection{Proprietary Discussion Forum Dataset}
\label{sec:proprietary_dataset}
Through an existing institutional partnership with a large public university in the United States, we retrieve student discussion forum posts in their deployed learning management system, covering 15 academic terms from Fall 2019 through Winter 2024. The data access is approved by the Institutional Review Board.

The full corpus comprises over four million posts across nearly two thousand courses. Most of these are non-question posts, featuring instructor-issued task prompts with limited stylistic variation; therefore, we restrict our analysis to student-originated academic questions, rather than more logistics-focused queries and announcements, to better capture diverse writing styles.

To identify these questions at scale, we use a two-stage filtering pipeline. We first retain only posts containing a question mark. From these, we randomly sample about 25,000 posts and prompt GPT-4o-mini to classify each as academic or not, yielding 10,074 academic questions. The filtered data includes 7000+ unique students, more than 50 departments, and 700+ courses. 

We generate the cognitive engagement level of forum posts based on Bloom's taxonomy \cite{krathwohl2002revision} using Qwen3-30B-A3B-Instruct-2507 \cite{yang2025qwen3}, following guidance from learning science co-authors and motivated by prior evidence of its applicability to related tasks \cite{wang2026memorization,badry2026cognitively}. In a random sample of 100 LLM-generated silver labels, 88\% match domain expert judgments. We formulate this as a multi-class classification task, assigning each post to one of six cognitive complexity levels. This tests whether synthesized posts preserve the cognitive and linguistic properties that distinguish higher-order thinking (e.g., analysis, evaluation) from lower-order thinking (e.g., memorization).

\begin{table}[!t]
\small
\centering\begin{tabular}{lccc}
\toprule
Dataset & Posts & Task \\
\midrule
Proprietary & 10,074 & Cognitive Engagement \\
Stanford MOOC & 29,604 &  Confusion\\
Stack Exchange & 10,000 &  Subject \\
\bottomrule
\end{tabular}
\caption{Dataset statistics and downstream task types.}
\label{tab:dataset_stats}
\end{table}

\begin{figure*}[!h]
    \centering
    \includegraphics[width=0.7\textwidth]{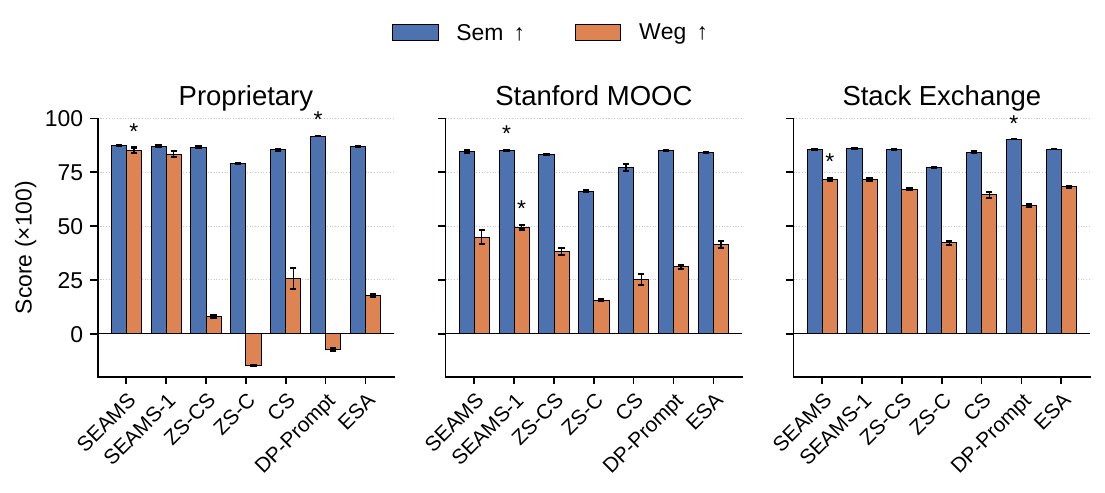}
    \caption{Reconstruction quality of our methods on three datasets, measured as semantic similarities and stylistic similarities in the embedding space of \citet{wegmann2022same}. The best-performing approaches are marked with *.}
    \label{fig:gen_quality_bars}
\end{figure*}

\subsection{Stanford MOOCPosts}
The Stanford MOOCPosts dataset \cite{agrawal2015youedu} contains 29,604 discussion forum posts from Stanford's online courses. We use pre-existing labels for confusion, treating them as linear regression labels with values ranging from 1 to 7.

\subsection{Stack Exchange}
We sample 10,000 posts evenly from Stack Exchange across 10 subject areas\footnote{\texttt{biology}, \texttt{chemistry}, \texttt{cs}, \texttt{cstheory}, \texttt{economics}, \texttt{linguistics}, \texttt{math}, \texttt{philosophy}, \texttt{physics}, \texttt{stats}}. The downstream task is subject classification: predicting which of the 10 sites a question was posted to. Subject is carried mainly by content, complementing Bloom's taxonomy level and expressed confusion.

\begin{figure*}
    \centering
    \includegraphics[width=0.8\textwidth]{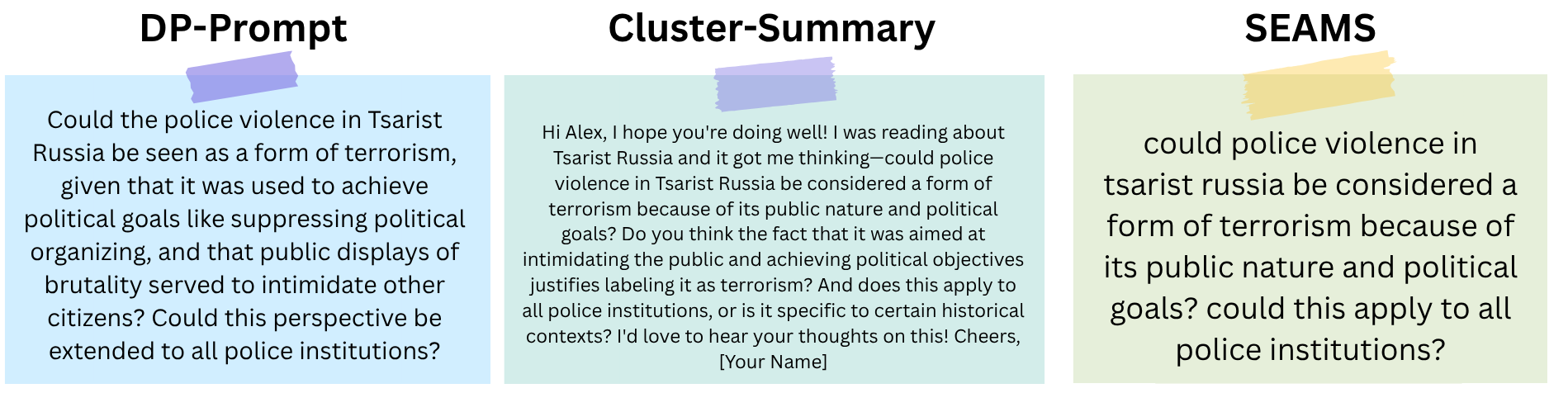}
    \caption{Example generations from DP-Prompt, Cluster-Summary-Centric, and SEAMS given the same input. Note that DP-Prompt capitalizes and formalizes the post, CS creates an email-like structure, while SEAMS achieves a register similar to typical student discussion posts.}
    \label{fig:examples}
\end{figure*}

\section{Experiments}
In our experiments, we set $K=20$ and $\mathcal{M}=5$ for SEAMS (aggregation settings in Appendix~\ref{app:aggregation}). We use \texttt{Qwen-2.5-72B-Instruct} to run all generations, unless otherwise specified. We use this model for its strong instruction-following and generation quality at a size that is feasible to serve locally over corpora of our scale. Appendix~\ref{app:compute} reports hardware, wall-clock, and token costs.

\subsection{Baselines}

\noindent \textbf{Naive Zero-shot Generation:} We examine whether the content and the structural channels alone are sufficient to reconstruct discussion forum posts. We study two baselines: \textbf{Zero-shot-Content (\textit{ZS-C})}, where the model is provided only the bullet summaries, and \textbf{Zero-shot-Content-Structure (\textit{ZS-CS})}, where both the bullet summaries and the extracted templates are supplied.

\noindent \textbf{Cluster-Summary-Centric (\textit{CS}):} Another natural way to utilize our cluster structures from SEAMS, but with enhanced privacy, is to condense clusters to their summaries. Summaries are generated by repeatedly sampling from the cluster and producing a series of summary candidates; the final summary is the one closest in the semantic embedding space to the averaged candidate embedding. This is similar to our aggregation mechanism. We then employ contrastive summaries to prevent all cluster summaries from becoming too similar, thereby preserving cluster-level signals (Appendix \ref{appendix:cluster-summary}).

The naive and summary-based baselines all leverage variants of the SEAMS prompt to ensure fair comparison. Instead of referring to the style context, these baseline prompts instruct the LLM to follow the tone and register typical of a student discussion forum post. 

\noindent \textbf{SEAMS with $\mathcal{M} = 1$ (\textit{SEAMS-1}):} To isolate the effect of aggregating across multiple candidate generations. For fair comparison, we use the same $K$; that is, each LM call for SEAMS has $K / \mathcal{M}$ posts, but each call for SEAMS-1 has $K$ posts. This corresponds to the standard retrieval-based in-context learning approach.

\noindent \textbf{\textit{DP-Prompt}:} Following \citet{utpala2023locally}'s main experimental setup, we run DP-Prompt without logit clipping. Clipping was reported in their work to provide a modest privacy-utility tradeoff, but is not used in their headline comparisons; we similarly omit it to faithfully reproduce the published method as evaluated. We swept across different temperatures and chose $t=1.0$ because it achieved the best reconstruction quality.

\noindent \textbf{Embedding Space Aggregation (\textit{ESA}):} \citet{wu2024privacy} produces the differentially-private output from candidates generated \textit{without} private demonstrations, choosing the one with the highest cosine similarity to the noisy aggregate embedding. For this baseline, we instantiate the candidates as ZS-CS examples, then select the output accordingly.

\subsection{Cluster Validation}

We inspect the 14 final meta-clusters on our proprietary dataset to judge their stylistic distinctiveness (Appendix \ref{app:cluster_analysis}). We discover various stylistic registers, from formulaic questions to analytical essays, supporting the use of clusters as style signals.


\subsection{Reconstruction Quality}

To improve robustness, we repeat all experiments across five seeds. We sample 1,000 validation posts and 1,000 test posts, stratified by post-content categories (course department or discussion topic). We evaluate generation quality along two dimensions: semantic preservation and style fidelity. Semantic preservation is measured using sentence embedding similarity from \texttt{all-mpnet-base-v2}; style fidelity is measured using style embedding similarity under both \texttt{StyleDistance} and \citet{wegmann2022same}, which is used to ensure evaluation robustness by providing an embedding separate from the SEAMS mechanisms.

Results are reported in Figure \ref{fig:gen_quality_bars}, with full results in Table \ref{tab:generation_quality}. Overall, we find that \textbf{SEAMS preserves stylistic similarity to the original student data across domains}. 
SEAMS-1 performs comparably to SEAMS on fidelity across all three datasets, and slightly better on Stanford MOOC, indicating that \textit{fidelity gains} come from in-context exemplar conditioning, less from aggregating across partitions. For the naive approaches, ZS-CS often outperforms ZS-C, suggesting that our structural templates are useful for generation. ESA, which selects among ZS-CS candidates, still falls below SEAMS because it selects from stylistically weak exemplars, and selection cannot compensate for missing style. Finally, most reconstruction methods achieve reasonable semantic similarity, suggesting that content and structural channels are generally sufficient to preserve meaning. In contrast, \textbf{achieving high style fidelity remains the primary bottleneck}.

\subsection{Downstream Performance}

As another measure of utility, we compare downstream evaluation metrics when models are trained on original vs. reconstructed data. We compute the difference between these metrics as a proxy for utility preservation. Since we formulate the Stanford MOOCPost downstream task as a regression problem, we use $R^2$ as the evaluation metric; for the other two tasks, we use macro-F1.

We fine-tune \texttt{distilbert-base-uncased} on the \emph{validation} split, using either the original posts or their rewrites constructed in the previous experiments, and evaluate on the original \textit{test} posts. Hyperparameters are fixed for all methods\footnote{3 epochs, learning rate $5\times10^{-5}$, batch size 16.}. 

\begin{table}[t]
  \centering
  \small
  \setlength{\tabcolsep}{3pt}
  \begin{tabular}{lccc}
    \toprule
    Method & Proprietary & Stanford & Stack Exch. \\
    \midrule
    SEAMS     & $-1.8 \pm 0.5$ & $\mathbf{-0.3 \pm 2.0}$ & $-0.1 \pm 0.6$ \\
    SEAMS-1   & $-2.5 \pm 3.4$ & $-1.0 \pm 1.9$ & $\mathbf{+0.0 \pm 1.0}$ \\
    ESA       & $-3.5 \pm 1.1$ & $-2.1 \pm 4.3$ & $+0.3 \pm 0.9$ \\
    CS        & $-4.8 \pm 1.0$ & $-9.3 \pm 3.0$ & $+0.2 \pm 1.2$ \\
    DP-Prompt & $\mathbf{-0.4 \pm 1.0}$ & $-1.1 \pm 1.2$ & $+0.1 \pm 0.6$ \\
    ZS-CS     & $-3.1 \pm 0.8$ & $-3.0 \pm 2.4$ & $+0.0 \pm 0.8$ \\
    ZS-C      & $-5.6 \pm 2.1$ & $-10.1 \pm 4.1$ & $-1.9 \pm 2.1$ \\
    \bottomrule
  \end{tabular}
  \caption{%
    Downstream utility of fine-tuned DistilBERT classifiers.
    $\Delta$ = score when training on rewrites $-$ score when training on the same original posts. Closer to zero is better, and the best methods are bolded.
  }
  \label{tab:downstream_utility}
\end{table}

Table~\ref{tab:downstream_utility} reports the difference in downstream metric values across datasets and reconstruction approaches. Classifiers trained on SEAMS rewrites lose at most 1.8 points relative to training on original posts, and SEAMS retains the most task signal on Stanford MOOC. Confusion (Stanford MOOC), which is expressed largely through affect and register, most clearly separates the methods, whereas subject (Stack Exchange) is carried by content: every method except ZS-C stays within 0.3 points of training on original posts. On Bloom's level, which depends more on content, DP-Prompt retains slightly more signal than SEAMS as a better semantic-preserving method. 
SEAMS also \textbf{preserves the within-cluster vs. between-cluster style separation} observed in the original data, indicating that population-level stylistic structure survives the rewrite (Appendix \ref{app:relational_structure}).


\begin{table}[t]
\centering
\small
\setlength{\tabcolsep}{3pt}
\begin{tabular}{lccc}
\toprule
\textbf{SEAMS vs.} & \textbf{Win rate} & \textbf{No boilerplate}\\
\midrule
ZS-CS     & $71.3 \pm 6.2$ & $65.7 \pm 8.2$ \\
CS        & $66.5 \pm 6.6$ & $61.7 \pm 7.6$ \\
DP-Prompt & $67.1 \pm 5.7$ & $67.7 \pm 5.6$ \\
\bottomrule
\end{tabular}
\caption{Percentage of human judgments in which the SEAMS rewrite was rated closer to the original style than the baseline rewrite, with item-bootstrap CI. 
}
\label{tab:style_match}
\end{table}

\begin{figure*}[t]
    \centering
    \begin{minipage}[b]{0.685\textwidth}
        \centering
        \includegraphics[width=\linewidth]{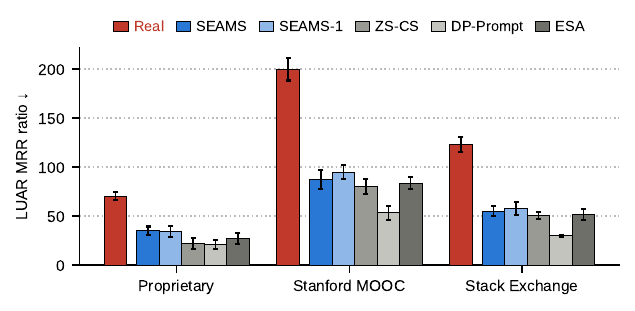}\\
        \small (a) LUAR attribution risk on three datasets
    \end{minipage}\hfill
    \begin{minipage}[b]{0.295\textwidth}
        \centering
        \includegraphics[width=\linewidth]{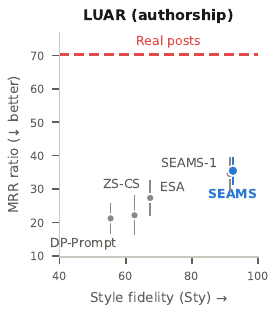}\\
        \small (b) Fidelity vs.\ LUAR risk (Proprietary)
    \end{minipage}
    \caption{(a) Attribution risk under LUAR when the attacker ranks all students: MRR ratio (MRR\,/\,chance-level MRR; 1.0 $=$ random attribution, lower is better), mean\,$\pm$\,std over 5 seeds.
    Results for all attackers, including a within-cluster analysis, are in Appendix~\ref{app:extra_attackers}. (b) Style fidelity vs.\ LUAR attribution risk on the proprietary dataset.}
    \label{fig:authorship}
\end{figure*}

\subsection{Human Evaluation}
\label{sec:human-eval}
We strengthen our results with a style-similarity judgment study on Prolific. Each one of 180 sampled items shows an original post from the proprietary dataset, paired with two randomly-ordered rewrites of that same post, one from SEAMS and one from a baseline (ZS-CS, CS, or DP-Prompt). Annotators choose which rewrite ``sounds more like it was written by the same person who wrote the original,'' focusing on wording and style rather than content, or judge them equally similar (Appendix~\ref{app:annotation}). Participants were current college or graduate students compensated at \$15 per hour; each rated 30 items plus three attention checks, and failing a check excluded them. Of 32 participants who started, 24 completed the study, yielding 3-5 ratings per item.

Annotators judged SEAMS rewrites closer to the original author's style than every baseline (Table~\ref{tab:style_match}). Because ZS-CS and CS outputs often contain boilerplates (e.g., ``[Your Name]'') that could cue annotators, we also report items where neither rewrite contains such boilerplates and find that SEAMS's advantage persists. Raters show fair agreement (Gwet's AC1 $= 0.303$; \citealp{gwet2008computing}), which we report as the primary statistic because SEAMS is preferred at a similar rate on nearly every item, a high-prevalence setting in which Krippendorff's $\alpha$ understates agreement ($\alpha = 0.103 \pm 0.075$); such levels are typical for subjective judgments of generated text \citep{clark2021all, karpinska2021perils, amidei2019agreement}. In an exploratory analysis, pairwise agreement increases with the style gap between the two rewrites (Spearman $\rho = 0.17$, $p < 0.05$, using \citet{wegmann2022same} embeddings), suggesting disagreement concentrates on items where the rewrites are stylistically close. 
 


\subsection{Authorship Attacks}
\label{sec:authorship}

Prior work on authorship obfuscation \citep{brennan2012adversarial, potthast2016author} often targets stylometric attackers,\footnote{An \emph{authorship attacker} is a model that, given a text and writing samples from a set of candidate authors, ranks the candidates by how likely each is to have written it. \emph{Stylometric} attackers rely on hand-crafted style features such as function-word frequencies, character $n$-grams, punctuation, and sentence length \citep{burrows2002delta}; \emph{embedding-based} attackers compare learned neural representations.} while we evaluate both stylometric and embedding-based ones. We evaluate on all three datasets; because Stack Exchange authors are public users, results there measure how much individual style survives rewriting rather than re-identification risk. Our central claim is that style-cluster conditioning pulls reconstructions toward the cluster mean, thereby reducing the individual authorship signal. An attacker holding other posts by every student in the training data tries to identify the author of a released rewrite, ranking all students; Appendix~\ref{app:extra_attackers} additionally restricts candidates to the author's style cluster.

We use three attackers. Two are retrieval-based and rank candidates by the maximum cosine similarity between the input and a candidate's posts: \citet{wegmann2022same} (Weg-ret) embeddings and LUAR \citep{uar-emnlp2021}, an authorship representation trained to identify authors across documents. The third is stylometric (Stylo-clf; \citealp{burrows2002delta}), a logistic regression classifier using the feature \textbf{difference} between the rewrite and a candidate post to predict same- vs.\ different-author. We report the \textbf{MRR ratio}: the true author's mean reciprocal rank divided by its chance level over the same posts. A ratio of $1.0$ indicates random attribution; higher values indicate more exploitable authorship signal.

SEAMS reduces MRR ratios relative to real posts for all three attackers on all three datasets (LUAR in Figure~\ref{fig:authorship}a; others in Appendix~\ref{app:extra_attackers}). The reduction is largest for the strongest attacker, LUAR (paired $t$-test over seeds, $p{<}0.001$ on each), evidence that \textbf{SEAMS empirically reduces attribution risk} relative to the original data. 
SEAMS is also below SEAMS-1 on six of nine attacker-dataset pairs, suggesting that the partition-then-aggregate approach contributes modestly to author protection.
Consistent with the well-known privacy-utility tradeoff, ZS-CS, ESA, and DP-Prompt achieve stronger MRR ratio reduction on the proprietary dataset, but sacrifice style fidelity (Figure~\ref{fig:authorship}b, Table~\ref{tab:generation_quality}). On Stanford MOOC and Stack Exchange, where style fidelity is similar across methods, all rewriting methods except DP-Prompt each similar attribution risk. We see that \textbf{SEAMS balances privacy and utility} by providing meaningful risk reduction across attack modes while maintaining downstream utility. 


\section{Discussion and Conclusion}

\noindent \textbf{Style as the Bottleneck} Across our experiments, reconstruction methods reliably preserve content but struggle to maintain an authentic student register. Even with high-level stylistic instructions, their rewrites often show limited linguistic diversity. The example generations (Figure \ref{fig:examples}) suggest that Qwen-2.5-72B has a weak prior understanding of student discussion style. The stronger style preservation of exemplar-conditioned methods across datasets indicates that for style-driven generation in non-mainstream registers, in-context exemplars convey information that instructions alone do not. This motivates using \textit{exemplar-based approaches} to synthesize data from non-mainstream domains. Refreshing the exemplar pool with SEAMS rewrites may further reduce attribution risk without a significant decline in fidelity, but we leave its systematic study to future work.

\noindent \textbf{Balancing Empirical Privacy and Utility} Prior differentially private text generation methods often incur significant utility costs, as illustrated by DP-Prompt's low stylistic similarity to the original data. SEAMS balances privacy and utility differently: it forgoes formal guarantees but achieves a moderate empirical reduction in within-cluster authorship attribution risk while maintaining reconstruction fidelity sufficient for downstream training. 




\noindent \textbf{Conclusion} In summary, we present SEAMS, a multi-channel, cluster-centric framework for high-fidelity, privacy-motivated reconstruction of student text. By isolating content, structure, and style into conditioning channels and aggregating across student-disjoint partitions within stylistic clusters, SEAMS \textit{preserves high style fidelity and downstream utility} on educationally meaningful tasks while \textit{empirically reducing authorship attribution risk}, a combination not achieved by prior methods we evaluated.

\section*{Limitation}

One limitation of our work is that we do not offer formal differential privacy guarantees. Our privacy claim is strictly empirical and is demonstrated through its ability to mitigate author attribution risk in Section~\ref{sec:authorship}. Practitioners facing requirements of formal $\epsilon$ guarantees should choose the appropriate DP methods.

We report ablations of cluster-restricted retrieval, noisy aggregation, and candidate selection in Appendix~\ref{sec:ablations}; their attack differences fall within seed variance, so the privacy contribution of individual components remains uncertain. Our SEAMS vs.\ SEAMS-1 comparison suggests partitioning contributes modestly to attack resistance.

SEAMS is not the most privacy-preserving method, as seen in the Authorship Attack section. While the tradeoff between privacy and utility is well-established, there is still room for improvement, potentially through performing generation with in-context examples generated through SEAMS to further reduce attribution risk, or using more computationally expensive approaches that offer more formal DP guarantees.


Inter-rater agreement in our human evaluation is low (ordinal Krippendorff's $\alpha = 0.103$; Gwet's AC1 $= 0.303$), reflecting the subjectivity of style judgments. We were able to attribute the low agreement potentially to participants disagreeing on stylistically similar rewrites of the same original posts, as there are irreducible variabilities in human language.

Finally, we only evaluate on one generator model, as our work aims to compare different methods using the same underlying model. However, it is possible that changing the model would alter the effectiveness of each of the methods studied. We leave testing the effect of varying the generator model to future work.

\section*{Ethical Considerations}

Our proprietary dataset is collected and analyzed in accordance with the existing IRB protocol. The data has not been anonymized at the source. All third-party models and datasets are used in accordance with their respective licenses.

For institutions planning to use SEAMS to rewrite their data for release, we note that SEAMS does not inherently support redacting personally identifiable information. Therefore, any rewritten data should undergo PII redaction or abstraction prior to release.

\bibliography{custom}

\appendix

\section{AI Assistant Usage}

In this project, in addition to using LLMs for experimentation purposes, we used AI assistants and agents to help write code and revise sections of the paper. 

\section{Embedding Model Selection}
\label{app:embedding_selection}

\subsection{Overview}


We compare four embedding models on their ability to separate same-student from different-student post pairs, specifically \texttt{StyleDistance} \cite{patel2025styledistance}, \citet{wegmann2022same}, Coh-Metrix \cite{graesser2004coh}, and \texttt{all-Mini-LM-v6} \cite{reimers2019sentence}. Discriminability is measured via the $t$-test of within- versus between-student pairwise cosine similarities, computed on 319 proprietary student forum posts drawn from 30 distinct students (10--17 posts per student). StyleDistance with PCA reduction to $d{=}128$ achieves the largest separation and is adopted for the per-seed style clustering.

\subsection{Embeddings Compared}

\begin{itemize}
  \item \textbf{Coh-Metrix} \citep{graesser2004coh}: 108 hand-crafted indices of text cohesion, syntactic complexity, and lexical sophistication. We drop 4 zero-variance indices and z-score-normalize the remaining 104 features.
  \item \textbf{StyleDistance} \citep{patel2025styledistance}: a 768-dimensional neural style embedding trained with synthetic parallel examples for content-independent style representation.
  \item \textbf{AnnaWegmann Style-Embedding} \citep{wegmann2022same}: a 768-dimensional neural style embedding trained with a contrastive same-author objective.
  \item \textbf{all-MiniLM-L6-v2} \citep{reimers2019sentence}: a 384-dimensional general-purpose semantic sentence embedding, included as a content-driven baseline.
\end{itemize}

For each neural embedding we additionally evaluate PCA-reduced variants with $d \in \{32, 64, 128, 256\}$. For Coh-Metrix we evaluate $d \in \{32, 64\}$.

\subsection{Evaluation Protocol}

For each embedding we compute pairwise cosine similarities between all 319 posts and partition the pairs into two distributions according to whether the two posts share the same author. The 30-student sample yields $1{,}567$ within-student pairs and $49{,}154$ between-student pairs. A discriminative style embedding produces a clear gap, with within-student similarities markedly higher than between-student similarities. We quantify the separation with an independent $t$-test.

\subsection{Results}
Table~\ref{tab:embedding-selection} reports the best PCA configuration per embedding. StyleDistance with $d{=}128$ achieves the largest $t$-statistic, outperforming the other neural style embedding AnnaWegmann, the semantic baseline all-MiniLM, and the hand-crafted Coh-Metrix features.

\begin{table}[!ht]
\centering
\small
\begin{tabular}{lccc}
\toprule
\textbf{Method} & \textbf{Best dim} & $t$-stat & within / between \\
\midrule
Coh-Metrix       & OG (104) & 25.52 & \phantom{-}0.146 / -0.005 \\
StyleDistance    & $d{=}128$ & \textbf{34.71} & \phantom{-}0.250 / -0.007 \\
AnnaWegmann      & $d{=}128$ & 26.01 & \phantom{-}0.243 / -0.002 \\
all-MiniLM-L6-v2 & $d{=}256$ & 30.33 & \phantom{-}0.107 / -0.006 \\
\bottomrule
\end{tabular}
\caption{Embedding model comparison on 319 posts from 30 students. ``Best dim'' is the configuration giving the largest $t$-statistic per method. For Coh-Metrix this is the original dimensionality. For the neural embeddings it is the best $d \in \{32, 64, 128, 256\}$ PCA variant.}
\label{tab:embedding-selection}
\end{table}

PCA consistently helps the neural embeddings: reducing StyleDistance from 768 to 128 dimensions raises its $t$-statistic from 20.70 to 34.71, and AnnaWegmann shows a similar jump (10.56 to 26.01). The raw outputs contain noisy dimensions that dilute the discriminative signal. Coh-Metrix, already low-dimensional, shows little change under PCA.

The relatively strong performance of all-MiniLM-L6-v2 likely reflects a content confound rather than true author differentiation. Students in our dataset tend to post within the same course, so within-student pairs are often within-course pairs, and topical similarity inflates within-student scores. StyleDistance is explicitly trained for content-independent style representation, and the fact that it still outperforms the semantic baseline suggests its advantage reflects a genuine stylistic signal rather than residual topical similarity.

\section{Bullet Summarizer Details}
\label{app:bullet_summarizer}

\subsection{Overview}

Each student forum post is summarized into bullet points using Qwen-2.5-7B-Instruct~\cite{qwen2024qwen2} served locally via vLLM. The prompt is optimized with GEPA~\cite{agrawal2025gepa}. The final artifact is the optimized prompt paired with fixed model weights; no student data touches the optimizer.

\subsection{Optimization Objective}

GEPA maximizes a scalar score computed over each (post, bullets) pair:

\begin{equation}
\label{eq:bullet_score}
\begin{aligned}
\mathcal{S}
= {}& w_{\text{sem}}\,\text{sim}
   + w_{\text{recon}}\,\text{ROUGE-1}_{\text{recon}} \\
  &- w_{\text{style}}\,\text{style\_sim}
   - w_{\text{len}}\,\ell_{\text{pen}} .
\end{aligned}
\end{equation}

\noindent where each term is defined as follows.

\paragraph{Semantic preservation ($\text{sim}$).}
Cosine similarity between the \texttt{all-mpnet-base-v2} embeddings of the post and its bullets. Rewards retaining the informational content of the post.

\paragraph{Stylistic dissimilarity ($\text{style\_sim}$).}
Cosine similarity between the \texttt{StyleDistance} embeddings of the post and its bullets. This term is \emph{subtracted}: lower style similarity means the bullets do not carry the author's writing style, which is the desired outcome. If bullets encode style, the cluster-ID conditioning applied downstream becomes redundant at inference time.

\paragraph{Reconstruction fidelity ($\text{ROUGE-1}_{\text{recon}}$).}
ROUGE-1 F\textsubscript{1} between the original post and a round-trip reconstruction of it from the bullets alone. A separate \texttt{Predict} module (same task LM) is prompted to reconstruct the original post from the bullets; poor reconstruction indicates the bullets omit key specifics.

\paragraph{Length penalty ($\ell_{\text{pen}}$).}
\begin{equation}
\begin{aligned}
\ell_{\text{pen}}
= {}& \min\!\left(
1,\;
\max\!\left(
0,\;
\frac{|\text{bullets}|_{\text{char}}}{|\text{post}|_{\text{char}}}
- \tau
\right)
\right), \\
& \tau = 0.30 .
\end{aligned}
\end{equation}
Zero when bullets are at most 30\% of the post length by character count; grows linearly beyond that threshold and is capped at 1.

\paragraph{Weights.} The default weights are $w_{\text{sem}}=1.0$, $w_{\text{recon}}=0.5$, $w_{\text{style}}=0.6$, $w_{\text{len}}=0.4$.

When GEPA's reflection pass is active, the metric also returns natural-language feedback to the reflection LM (GPT-5) to guide prompt revision.

\subsection{Training Data}

GEPA is run on Stack Exchange posts (200 training, 150 validation examples), drawn from a filtered CSV requiring $\geq 100$ characters per post. Student data is never used during optimization.

\subsection{In-Context Demonstrations}

Eight manually constructed (post, bullets) demonstrations are provided at inference time to cover distinct failure modes identified in early spot-checks.

\section{Academic-Content-Removed Template Details}
\label{app:rr_qa}

\subsection{Overview}

Each student post is converted into a \emph{template} in which domain-specific academic content is replaced by one of four bracketed markers, leaving all non-domain text (connectives, hedging, filler, opinion markers, punctuation) intact. The templating prompt is optimized with GEPA~\cite{agrawal2025gepa} against expert human annotations on 50 proprietary student posts, maximizing character-level agreement with the human reference masks. The task LM is GPT-5-mini; the reflection LM is GPT-5.

\subsection{Marker Vocabulary}

Four markers cover the domain-content categories identified by our education collaborators:

\begin{center}
\begin{tabular}{lp{5cm}}
\toprule
\textbf{Marker} & \textbf{Semantics} \\
\midrule
\texttt{\#\#\#} & Named concept, theory, formula, or domain term the student is asking about \\
\texttt{***}    & The question body directly related to the target concept (what is explicitly being asked) \\
\texttt{\%\%\%} & The student's explanation of their own understanding of the concept \\
\texttt{\textasciicircum\textasciicircum\textasciicircum} & Context or condition of the question (a specific scenario, given information, or described situation) \\
\bottomrule
\end{tabular}
\end{center}

Two rules govern the output: (1) each masked span must start and end exactly at the boundaries of the domain-specific content; (2) if two consecutive sentences belong to the same marker type, they are separated by a period rather than concatenated (e.g.\ \texttt{\%\%\%.\%\%\%} not \texttt{\%\%\%\%\%\%}).

During SEAMS, we then perform the replacement of markers into free texts so that SEAMS has better context.

\subsection{Optimization Objective}

\paragraph{Character-level accuracy.}
Given an original post $p$, a model-produced template $\hat{r}$, and a human-annotated reference template $r^*$, the score is the fraction of character positions on which the model and human agree about whether that character is masked:

\begin{equation}
\label{eq:char_acc}
\begin{aligned}
&\text{char\_acc}(p, \hat{r}, r^*) \\
= {}& \frac{1}{|p|}
\sum_{i=1}^{|p|}
\mathbf{1}\!\left[
\text{masked}_{\hat{r}}(i) 
=
\text{masked}_{r^*}(i)
\right].
\end{aligned}
\end{equation}

\noindent where $\text{masked}_r(i) = \mathbf{1}[\text{character } i \text{ of } p \text{ is replaced by a marker in } r]$, determined by splitting $r$ on the marker regex \texttt{[\textbackslash*\#\textasciicircum\%]+} and locating the surviving literal substrings in $p$. The metric also returns natural-language feedback to guide GEPA's reflection pass.

\subsection{Training Data}

Optimization uses 50 expert-annotated student posts (40 train / 50 validation, all available examples used as the validation set). The annotations were produced by education collaborators and cover posts from multiple course departments. Because this data is proprietary, the released artifact is the optimized prompt only; no student post text is embedded in the model weights.

\subsection{In-Context Demonstrations}

Six manually constructed (post, template) demonstrations are provided at inference time, selected to cover recurring structural patterns. 

\section{MIPROv2 Prompt Proposal for the Rewriter}
\label{app:mipro}

\subsection{Instruction Candidate Proposal}

We use MIPROv2~\cite{opsahl2024optimizing} in proposal-only mode: we call \texttt{\_propose\_instructions} to generate $C = 10$ candidate instruction strings without any few-shot bootstrapping (\texttt{max\_bootstrapped\_demos} $= 0$, \texttt{max\_labeled\_demos} $= 0$).

\subsection{Candidate Evaluation}

Each candidate instruction is evaluated on a sample of $n = 100$ held-out posts drawn from the validation split. For each post the rewriter is called with the top-$K$ ($K=20$) most style-similar training posts as exemplars, retrieved by cosine similarity under \texttt{StyleDistance}~\cite{patel2025styledistance} embeddings. The per-example score used to rank candidates is:

\begin{equation}
\label{eq:mipro_metric}
s = \tfrac{1}{3}\,\text{style\_sim} + \tfrac{1}{3}\,\text{sem\_sim} + \tfrac{1}{3}\,\text{ROUGE-1}_{F_1}
\end{equation}

\noindent where $\text{style\_sim}$ is cosine similarity between \texttt{StyleDistance} embeddings of the generated and original post, $\text{sem\_sim}$ is cosine similarity between \texttt{all-mpnet-base-v2} embeddings, and $\text{ROUGE-1}_{F_1}$ is unigram overlap with the original post.

Candidates are ranked by mean $\text{sem\_sim}$ across the evaluation sample; the top-ranked instruction is adopted as the fixed rewriter prompt for all downstream runs.

\section{Cluster Summary Generation}
\label{appendix:cluster-summary}

Each cluster is summarized via a two-stage process: \emph{contrastive prompt construction} followed by \emph{ensemble candidate selection}.

\paragraph{Focal and contrast post selection.}
For a given meta-cluster $c$, we first compute its meta-centroid $\bar{\mathbf{e}}_c$ as the $\ell_2$-normalized mean of the $\ell_2$-normalized cluster-center embeddings of all constituent seed clusters. We then collect the union of all post texts belonging to those seed clusters (deduplicating across seeds) and select the $K_{\text{focal}} = 15$ posts whose embeddings lie closest to $\bar{\mathbf{e}}_c$ in cosine similarity as the \emph{focal} set. To form the \emph{contrast} set, we randomly sample $K_{\text{contrast}} = 5$ posts from the union of all other meta-clusters, excluding posts that appear in $c$.

\paragraph{Axis-specific contrastive prompt.}
The LLM receives the focal and contrast posts and is instructed to produce a two-sentence summary. The first sentence describes what the focal posts share with student discussion posts \emph{in general} (i.e., the baseline register or content). The second sentence articulates what specifically distinguishes the focal cluster \emph{from} the contrast examples. This contrastive framing is critical for preventing all cluster summaries from collapsing to generic descriptions of student posts.

\paragraph{Ensemble candidate selection.}
We generate $N = 10$ candidate summaries by querying the LLM. Each candidate is embedded with a text embedding model (\texttt{text-embedding-3-small}), and the embeddings are $\ell_2$-normalized. The final summary is the candidate whose normalized embedding has the highest cosine similarity to the mean of all candidate embeddings:
\[
  s^* = \arg\max_{s_i} \; \hat{\mathbf{e}}_i^\top \bar{\mathbf{e}},
  \quad \bar{\mathbf{e}} = \frac{1}{N}\sum_{j=1}^{N} \hat{\mathbf{e}}_j,
\]
where $\hat{\mathbf{e}}_i$ is the normalized embedding of candidate $s_i$. This is the same centroid-proximity aggregation used in the privacy-motivated generation step, and it favors summaries that are semantically representative of the consensus across candidates while filtering out outlier generations.

\section{Cluster Inspection}
\label{app:cluster_analysis}
Table \ref{tab:cluster_summaries} presents the concise summaries of the 14 identified clusters, highlighting the most distinguishing features of each cluster. To further examine their conceptual validity, we reviewed the cluster content in relation to educational theories on student question-asking and academic help-seeking. Overall, the clusters align with prior research in learning science that students' questions could serve different functions, including clarifying confusion, checking understanding, asking for evidence, extending or elaborating ideas, or promoting peer discussion \cite{graesser1994question,karabenick2003seeking,king1994guiding}.

Specifically, the identified clusters reflect several theoretically meaningful patterns. Several capture \textbf{clarification-oriented help-seeking}, in which students pose short, task-focused questions to resolve confusion. For example, students ask questions (rewrite version) such as \textit{“Hi class, I got so confused with the last question on the weekly quiz. How does 1.98 come out in this question?”} Another set of clusters captures \textbf{peer-oriented interaction}, where students combine acknowledgment, feedback, and follow-up questions to sustain discussion with classmates. Representative example like \textit{“Hi xx, I like and definitely agree with your post. Here are some further questions to consider: How much do you think the public has become significantly better at openly stopping hate in person, so as to motivate them to mask spreading hate behind social media? ..."} In addition, a set of clusters reflects more \textbf{elaborate, reflective, and synthesis-oriented questions}. Those questions often summarize students' personal understanding and then connect it to broader topical discussion. For example, \textit{“In reading ..., I learned that there are some ethical theories holding that there exist universal ethical rules. In my understanding, when they say these rules are universal, they mean "they are universal for every human being in every possible occasion". However, isn't it a meaningless claim?” }Such questions move beyond clarification and invite interpretation, evaluation, and synthesis, aligning with the constructive engagement emphasized in the ICAP framework \cite{chi2014icap}. 

Overall, this inspection indicates that these clusters capture theoretically meaningful variations in students' question asking in discussion forums.

\begin{table}[!ht]
  \centering
  
  \small
  \setlength{\tabcolsep}{5pt}
  \begin{tabular}{l ccc}
    \toprule
    \textbf{Method} & \textbf{Semantic} & \textbf{Style} & \textbf{ROUGE-1} \\
    \midrule
    SEAMS$_{seed}$
      & 0.874 $\pm$ .004 & 0.923 $\pm$ .006 & 0.648 $\pm$ .006 \\
    SEAMS
      & 0.874 $\pm$ .004 & 0.923 $\pm$ .004 & 0.648 $\pm$ .005 \\
      \midrule
    CS$_{seed}$
      & 0.864 $\pm$ .004 & 0.608 $\pm$ .005 & 0.577 $\pm$ .010 \\
    CS
      & 0.854 $\pm$ .004 & \textbf{0.688 $\pm$ .016} & 0.549 $\pm$ .012 \\
    \bottomrule
  \end{tabular}
  \caption{%
    Ablation of per-seed vs.\ meta-aggregation ($n{=}1000$, averaged over 5 seeds).
    Values are mean $\pm$ std over seeds.
  }
  \label{tab:ablation_meta_vs_per_seed}
\end{table}

\begin{table*}[!t]
\centering
\footnotesize
\caption{Cluster Summary: Key Distinguishing Features}
\label{tab:cluster_summaries}
\begin{tabular}{p{0.1\textwidth} p{0.9\textwidth}}
\hline
\textbf{Cluster} & \textbf{Summarization} \\
\hline

1 & Dialogic peer responses that combine salutations, personalized feedback, reflective interpretation, and multiple probing follow-up questions. These posts often link course texts to broader implications. \\

2 & Strongly interpersonal peer replies that begin with greetings and praise, directly address classmates, and ask several conversational questions for clarification, reflection, or opinion. \\

3 & Elaborative peer-feedback posts that combine praise, constructive critique, personal reflection, and open-ended questions. They foreground relational engagement and continued dialogue. \\

4 & Relational and affective replies that acknowledge peers, connect to readings or personal experiences, and close with targeted reflective questions. \\

5 & Highly dialogic and informal exchanges that combine empathy, tentative suggestions, personal reactions, and exploratory clarification questions directed to peers or instructors. \\

6 & Mini-essay style posts that summarize and analyze readings or films, often using quotations, citations, and comparative or theoretical arguments to provoke discussion. \\

7 & Synthesis-oriented posts that interweave textual summary, interpretation, evaluative stance, and structured prompts for group work or class discussion. \\

8 & Concise, task-focused entries formatted as two numbered standalone content questions, with minimal greeting, first-person framing, or contextual elaboration. \\

9 & Formulaic peer-feedback replies that begin with a greeting and positive acknowledgment, followed by exactly two numbered follow-up questions. \\

10 & Structured responses to assigned prompts, often using labels or numbered answers, multi-paragraph synthesis, course readings, citations, and reflective questions. \\

11 & Terse, decontextualized technical or concept-clarification questions, often using domain-specific terms, acronyms, and stacked sub-questions with little interpersonal framing. \\

12 & Brief, polite help-seeking messages with greetings, explicit confusion or help-seeking phrases, and a single focused clarifying question. \\

13 & Casual, peer-directed dialogic posts with named addressees, brief references to readings or lectures, and dense strings of quick follow-up questions. \\

14 & Formulaic troubleshooting requests that open with greetings and gratitude or confusion markers, then ask narrowly scoped procedural or assignment-related questions. \\

\hline
\end{tabular}
\end{table*}

\section{Generation Quality: Full Table}

We present the full table in Table \ref{tab:generation_quality}. Lexical overlap is measured using ROUGE-1 \cite{lin2004rouge}. We additionally report diversity metrics, including MATTR \cite{covington2010cutting} and Self-BLEU \cite{zhu2018texygen}.

\begin{table*}[t]
\centering
\small
\begin{tabular}{llcccccc}
\toprule
\textbf{Dataset} & \textbf{Method} & \textbf{Sem} $\uparrow$ & \textbf{Sty} $\uparrow$ & \textbf{Weg} $\uparrow$ & \textbf{R-1} $\uparrow$ & \textbf{MATTR} & \textbf{Self-BLEU} $\downarrow$ \\
\midrule
\multirow{7}{*}{\textsc{Proprietary}}
 & SEAMS             & 87.4 $\pm$ 0.4 & \textbf{92.3 $\pm$ 0.4} & \textbf{85.3 $\pm$ 1.2} & \textbf{64.8 $\pm$ 0.5} & 85.2 & 19.8 \\
 & SEAMS-1           & 87.1 $\pm$ 0.4 & 91.4 $\pm$ 0.5 & 83.3 $\pm$ 1.4 & 63.8 $\pm$ 0.5 & 84.7 & 19.5 \\
 & ESA               & 87.0 $\pm$ 0.3 & 67.4 $\pm$ 0.2 & 17.7 $\pm$ 0.5 & 59.8 $\pm$ 0.3 & 85.3 & 23.0 \\
 & CS                & 85.4 $\pm$ 0.4 & 68.8 $\pm$ 1.6 & 25.7 $\pm$ 4.9 & 54.9 $\pm$ 1.2 & 86.0 & 27.7 \\
 & DP-Prompt (t=1.0) & \textbf{91.8 $\pm$ 0.1} & 55.4 $\pm$ 0.1 & $-7.3 \pm 0.5$ & 61.4 $\pm$ 0.2 & 86.7 & 15.6 \\
 & ZS-CS             & 86.6 $\pm$ 0.4 & 62.7 $\pm$ 0.3 & 7.9 $\pm$ 0.6 & 58.3 $\pm$ 0.3 & 85.1 & 23.5 \\
 & ZS-C              & 79.0 $\pm$ 0.2 & 55.2 $\pm$ 0.1 & $-14.6 \pm 0.2$ & 36.2 $\pm$ 0.2 & 86.4 & 37.7 \\
\midrule
\multirow{7}{*}{\textsc{Stanford MOOC}}
 & SEAMS             & 84.6 $\pm$ 0.7 & 82.3 $\pm$ 1.1 & 44.9 $\pm$ 3.1 & 68.7 $\pm$ 1.2 & 87.6 & 15.8 \\
 & SEAMS-1           & \textbf{85.2 $\pm$ 0.2} & \textbf{83.7 $\pm$ 0.4} & \textbf{49.5 $\pm$ 1.1} & \textbf{69.7 $\pm$ 0.4} & 87.5 & 17.0 \\
 & ESA               & 84.2 $\pm$ 0.2 & 81.4 $\pm$ 0.5 & 41.3 $\pm$ 1.7 & 66.9 $\pm$ 0.5 & 87.7 & 20.0 \\
 & CS                & 77.1 $\pm$ 1.6 & 76.0 $\pm$ 0.9 & 25.3 $\pm$ 2.5 & 51.2 $\pm$ 3.6 & 88.1 & 30.8 \\
 & DP-Prompt (t=1.0) & 85.0 $\pm$ 0.3 & 77.5 $\pm$ 0.3 & 31.1 $\pm$ 1.0 & 56.6 $\pm$ 0.4 & 89.3 & 13.4 \\
 & ZS-CS             & 83.3 $\pm$ 0.2 & 80.1 $\pm$ 0.5 & 38.1 $\pm$ 1.5 & 64.6 $\pm$ 0.5 & 87.5 & 20.5 \\
 & ZS-C              & 66.2 $\pm$ 0.4 & 74.4 $\pm$ 0.3 & 15.6 $\pm$ 0.6 & 28.7 $\pm$ 0.2 & 87.2 & 38.6 \\
\midrule
\multirow{7}{*}{\textsc{Stack Exchange}}
 & SEAMS             & 85.7 $\pm$ 0.3 & \textbf{92.0 $\pm$ 0.3} & \textbf{71.7 $\pm$ 0.6} & 63.4 $\pm$ 0.7 & 81.1 & 17.3 \\
 & SEAMS-1           & 86.0 $\pm$ 0.4 & 91.5 $\pm$ 0.2 & 71.6 $\pm$ 0.7 & \textbf{63.7 $\pm$ 0.3} & 81.5 & 16.5 \\
 & ESA               & 85.7 $\pm$ 0.1 & 90.7 $\pm$ 0.1 & 68.2 $\pm$ 0.4 & 62.1 $\pm$ 0.2 & 81.5 & 19.1 \\
 & CS                & 84.4 $\pm$ 0.6 & 88.9 $\pm$ 1.0 & 64.5 $\pm$ 1.5 & 54.0 $\pm$ 4.9 & 81.8 & 26.0 \\
 & DP-Prompt (t=1.0) & \textbf{90.3 $\pm$ 0.0} & 88.0 $\pm$ 0.2 & 59.5 $\pm$ 0.7 & 60.8 $\pm$ 0.2 & 82.0 & 18.1 \\
 & ZS-CS             & 85.5 $\pm$ 0.2 & 90.2 $\pm$ 0.1 & 67.2 $\pm$ 0.4 & 61.4 $\pm$ 0.2 & 81.5 & 19.0 \\
 & ZS-C              & 77.2 $\pm$ 0.3 & 82.6 $\pm$ 0.3 & 42.2 $\pm$ 1.0 & 35.7 $\pm$ 0.2 & 83.5 & 37.0 \\
\bottomrule
\end{tabular}
\caption{Generation quality across all three datasets. All values multiplied by 100; best per dataset-column in bold for Sem/Sty/Weg/R-1. \textsc{Sem} and \textsc{Sty} are cosine similarities to the original post in \texttt{all-mpnet-base-v2} and \texttt{StyleDistance/styledistance}; \textsc{Weg} is style similarity under \citet{wegmann2022same}, an embedder not used for retrieval. R-1 is ROUGE-1 F1 (stemmed). MATTR (window 50) and Self-BLEU measure lexical and generation diversity.}
\label{tab:generation_quality}
\end{table*}

\section{Relational Structure Preservation}
\label{app:relational_structure}

Beyond per-post fidelity, we ask whether SEAMS preserves \textit{population-level} stylistic structure, measured by the within-vs-between cluster style gap:
$\Delta\text{sep} = \overline{\cos}_{\text{within}} - \overline{\cos}_{\text{between}}$, 
the difference between mean pair-wise cosine similarity of within-cluster and between-cluster posts. Values close to the original indicate that population-level style survives the rewrite. We compute $\Delta\text{sep}$ in two embedding spaces to rule out artifacts of the space SEAMS uses for aggregation (Table~\ref{tab:relational_structure}).

\textbf{SEAMS reproduces the original $\Delta\text{sep}$ }in both spaces; SEAMS-1 behaves comparably, suggesting partitioning is not the source of structural preservation. CS substantially inflates $\Delta\text{sep}$, likely due to within-cluster homogenization from rewriting all members against a single cluster summary. This is a failure mode SEAMS avoids by conditioning on individual exemplars.

\begin{table}[t]
  \centering
  \small
  \begin{tabular}{@{}lcc@{}}
    \toprule
    Method & Within $\uparrow$ & $\Delta$sep \\
    \midrule
    \multicolumn{3}{l}{\textit{StyleDistance}} \\
    Original posts & $90.8 \pm 0.4$ & $6.5 \pm 0.3$ \\
    SEAMS  & $79.1 \pm 1.5$ & $\bm{8.3 \pm 0.6}$ \\
    SEAMS-1  & $76.5 \pm 1.4$ & $8.5 \pm 0.9$ \\
    CS  & $63.0 \pm 4.8$ & $33.9 \pm 5.4$ \\
    \midrule
    \multicolumn{3}{l}{\textit{Wegmann}} \\
    Original posts & $93.7 \pm 0.2$ & $5.5 \pm 0.2$ \\
    SEAMS  & $88.1 \pm 0.5$ & $\bm{5.0 \pm 0.4}$ \\
    SEAMS-1  & $86.6 \pm 0.5$ & $4.9 \pm 0.5$ \\
    CS  & $86.5 \pm 1.3$ & $13.2 \pm 2.1$ \\
    \bottomrule
  \end{tabular}
  \caption{Within-cluster embedding similarities and the difference between within- and between-cluster similarities, computed in both \texttt{StyleDistance} and \texttt{AnnaWegmann} style embedding space, across methods.}
  \label{tab:relational_structure}
\end{table}

\section{Authorship Attack: Full Table}
\label{app:extra_attackers}

We report two attack settings. In the \emph{global} setting used in Section~\ref{sec:authorship}, the attacker ranks every student with training posts, without knowing the true author's style cluster (Table~\ref{tab:authorship_global}). In the \emph{within-cluster} setting, candidates are restricted to the students in the true author's home cluster, the meta-cluster containing most of their training posts (Table~\ref{tab:authorship_full}). This removes easy negatives from other clusters and isolates the individual style that remains once cluster-level style is preserved; it also adds two further attackers: a Wegmann classifier (Weg-clf; logistic regression on the difference of \citet{wegmann2022same} embeddings) and a semantic retriever (MPNet-ret; \texttt{all-mpnet-base-v2}). MPNet-ret links posts through shared course content rather than style; since every faithful rewrite preserves content, no method reduces it substantially. In the within-cluster setting, SEAMS is below SEAMS-1 on seven of nine attacker--dataset pairs for the three main attackers.

\begin{table}[!h]
\centering
\footnotesize
\setlength{\tabcolsep}{3pt}
\begin{tabular}{lccc}
\toprule
\textbf{Method} & \textbf{Weg-ret} & \textbf{Stylo-clf} & \textbf{LUAR} \\
\midrule
\multicolumn{4}{l}{\textsc{Proprietary} (6{,}231 candidates)} \\
\quad Real (unmodified) & 10.18 $\pm$ 3.28 & 6.21 $\pm$ 0.61 & 70.28 $\pm$ 4.09 \\
\quad SEAMS & 7.88 $\pm$ 2.76 & 5.86 $\pm$ 0.79 & 35.47 $\pm$ 4.12 \\
\quad SEAMS-1 & 6.31 $\pm$ 2.38 & 6.48 $\pm$ 1.08 & 34.53 $\pm$ 5.21 \\
\quad ESA & 5.26 $\pm$ 1.13 & 4.31 $\pm$ 0.50 & 27.33 $\pm$ 5.25 \\
\quad ZS-CS & 4.04 $\pm$ 1.73 & 4.10 $\pm$ 0.87 & 22.15 $\pm$ 5.93 \\
\quad DP-Prompt & 2.62 $\pm$ 0.83 & 4.11 $\pm$ 1.08 & 21.22 $\pm$ 4.54 \\
\midrule
\multicolumn{4}{l}{\textsc{Stanford MOOC} (10{,}626 candidates)} \\
\quad Real (unmodified) & 88.05 $\pm$ 5.51 & 15.28 $\pm$ 3.72 & 200.06 $\pm$ 11.53 \\
\quad SEAMS & 35.65 $\pm$ 4.09 & 11.39 $\pm$ 4.19 & 87.55 $\pm$ 9.25 \\
\quad SEAMS-1 & 36.22 $\pm$ 2.37 & 11.82 $\pm$ 3.91 & 94.78 $\pm$ 7.17 \\
\quad ESA & 32.61 $\pm$ 3.21 & 11.14 $\pm$ 4.47 & 83.76 $\pm$ 5.95 \\
\quad ZS-CS & 30.67 $\pm$ 1.62 & 11.73 $\pm$ 4.73 & 80.50 $\pm$ 7.74 \\
\quad DP-Prompt & 27.60 $\pm$ 3.45 & 10.46 $\pm$ 3.09 & 53.61 $\pm$ 7.31 \\
\midrule
\multicolumn{4}{l}{\textsc{Stack Exchange} (5{,}702 candidates)} \\
\quad Real (unmodified) & 22.90 $\pm$ 4.93 & 31.39 $\pm$ 8.92 & 123.52 $\pm$ 7.79 \\
\quad SEAMS & 9.50 $\pm$ 1.89 & 12.36 $\pm$ 3.53 & 55.35 $\pm$ 5.49 \\
\quad SEAMS-1 & 10.41 $\pm$ 2.22 & 10.06 $\pm$ 4.57 & 57.81 $\pm$ 6.39 \\
\quad ESA & 8.18 $\pm$ 1.81 & 11.87 $\pm$ 4.40 & 51.85 $\pm$ 5.87 \\
\quad ZS-CS & 8.27 $\pm$ 0.81 & 12.11 $\pm$ 3.35 & 50.77 $\pm$ 3.46 \\
\quad DP-Prompt & 4.26 $\pm$ 0.69 & 9.61 $\pm$ 3.30 & 30.04 $\pm$ 1.00 \\
\bottomrule
\end{tabular}
\caption{Global authorship attack: the attacker ranks all students with training posts (MRR ratio, mean $\pm$ std over 5 seeds; 1.0 $=$ random chance, higher $=$ easier to attribute). This is the setting reported in Section~\ref{sec:authorship}.}
\label{tab:authorship_global}
\end{table}

\begin{table*}[!h]
\centering
\small
\setlength{\tabcolsep}{5pt}
\begin{tabular}{lccccc}
\toprule
\textbf{Method} & \textbf{Weg-ret} & \textbf{Stylo-clf} & \textbf{LUAR} & \textbf{Weg-clf} & \textbf{MPNet-ret} \\
\midrule
\multicolumn{6}{l}{\textsc{Proprietary}} \\
\quad Real (unmodified) & 2.30 $\pm$ 0.27 & 2.69 $\pm$ 0.39 & 8.80 $\pm$ 0.74 & 1.76 $\pm$ 0.19 & 6.53 $\pm$ 0.64 \\
\quad SEAMS & 1.89 $\pm$ 0.31 & 2.38 $\pm$ 0.31 & 5.56 $\pm$ 0.73 & 1.35 $\pm$ 0.20 & 5.86 $\pm$ 0.72 \\
\quad SEAMS-1 & 2.13 $\pm$ 0.28 & 2.37 $\pm$ 0.28 & 5.61 $\pm$ 0.83 & 1.49 $\pm$ 0.32 & 6.18 $\pm$ 0.72 \\
\quad ESA & 1.47 $\pm$ 0.37 & 1.91 $\pm$ 0.23 & 4.62 $\pm$ 0.92 & 1.09 $\pm$ 0.32 & 5.80 $\pm$ 0.70 \\
\quad ZS-CS & 1.46 $\pm$ 0.44 & 1.98 $\pm$ 0.11 & 3.94 $\pm$ 0.92 & 0.98 $\pm$ 0.26 & 5.79 $\pm$ 0.76 \\
\quad DP-Prompt & 1.34 $\pm$ 0.17 & 2.11 $\pm$ 0.37 & 4.45 $\pm$ 0.69 & 1.05 $\pm$ 0.23 & 6.29 $\pm$ 0.81 \\
\midrule
\multicolumn{6}{l}{\textsc{Stanford MOOC}} \\
\quad Real (unmodified) & 6.89 $\pm$ 0.21 & 4.36 $\pm$ 0.39 & 13.30 $\pm$ 1.13 & 4.73 $\pm$ 0.73 & 9.45 $\pm$ 0.66 \\
\quad SEAMS & 4.20 $\pm$ 0.25 & 4.00 $\pm$ 0.54 & 8.51 $\pm$ 0.37 & 3.55 $\pm$ 0.24 & 9.15 $\pm$ 0.40 \\
\quad SEAMS-1 & 4.42 $\pm$ 0.19 & 4.17 $\pm$ 0.47 & 8.90 $\pm$ 0.28 & 3.50 $\pm$ 0.38 & 9.09 $\pm$ 0.48 \\
\quad ESA & 4.05 $\pm$ 0.39 & 4.12 $\pm$ 0.72 & 8.46 $\pm$ 0.55 & 3.34 $\pm$ 0.27 & 9.00 $\pm$ 0.64 \\
\quad ZS-CS & 4.04 $\pm$ 0.17 & 4.18 $\pm$ 0.72 & 8.34 $\pm$ 0.66 & 3.61 $\pm$ 0.25 & 8.77 $\pm$ 0.65 \\
\quad DP-Prompt & 3.76 $\pm$ 0.51 & 3.64 $\pm$ 0.60 & 7.26 $\pm$ 0.33 & 3.03 $\pm$ 0.32 & 8.90 $\pm$ 0.48 \\
\midrule
\multicolumn{6}{l}{\textsc{Stack Exchange}} \\
\quad Real (unmodified) & 3.30 $\pm$ 0.28 & 5.38 $\pm$ 0.90 & 12.48 $\pm$ 1.67 & 2.30 $\pm$ 0.37 & 14.92 $\pm$ 0.73 \\
\quad SEAMS & 2.07 $\pm$ 0.19 & 3.40 $\pm$ 0.59 & 7.27 $\pm$ 0.75 & 1.70 $\pm$ 0.32 & 14.20 $\pm$ 0.61 \\
\quad SEAMS-1 & 2.32 $\pm$ 0.23 & 3.20 $\pm$ 0.47 & 7.95 $\pm$ 0.94 & 1.89 $\pm$ 0.31 & 14.13 $\pm$ 0.97 \\
\quad ESA & 1.89 $\pm$ 0.12 & 3.44 $\pm$ 0.43 & 6.99 $\pm$ 0.67 & 1.63 $\pm$ 0.23 & 14.28 $\pm$ 0.74 \\
\quad ZS-CS & 1.91 $\pm$ 0.14 & 3.26 $\pm$ 0.43 & 6.86 $\pm$ 0.85 & 1.52 $\pm$ 0.19 & 14.21 $\pm$ 0.78 \\
\quad DP-Prompt & 1.54 $\pm$ 0.21 & 3.06 $\pm$ 0.53 & 5.44 $\pm$ 0.71 & 1.58 $\pm$ 0.20 & 14.86 $\pm$ 0.50 \\
\bottomrule
\end{tabular}
\caption{Within-cluster authorship attack: candidates are the students in the true author's home cluster (MRR ratio, mean $\pm$ std over 5 seeds; 1.0 $=$ random chance, higher $=$ easier to attribute; protocol in Section~\ref{sec:authorship}).}
\label{tab:authorship_full}
\end{table*}

\newpage
\section{Ablation: Meta-Clustering Necessity}
\label{app:ablation_meta_cluster}

SEAMS uses meta-clusters formed by cross-seed agglomerative clustering rather than per-seed HDBSCAN clusters directly. We ablate this choice by comparing two variants of each method: \textbf{$X_{seed}$} variants retrieve same-cluster exemplars (SEAMS) or generate summaries (CS) using only the per-seed clusters from the current seed, while the other versions use the cross-seed meta-clusters.

Table~\ref{tab:ablation_meta_vs_per_seed} reports the results. For SEAMS, per-seed and meta-cluster variants are statistically indistinguishable on all three metrics, indicating that the meta-clustering step is not critical for high-fidelity exemplar-based generation. For CS, however, meta-clustering substantially improves style fidelity (0.688 vs.\ 0.608), at a small cost to semantic similarity and ROUGE-1.

This asymmetry is consistent with the failure mode of the CS pipeline. Per-seed clusters are smaller and more variable across seeds, so per-seed cluster summaries are noisier and less stylistically distinctive. Cross-seed aggregation stabilizes the cluster definitions, producing summaries that more consistently characterize the cluster's register. SEAMS does not depend on summary stability because it conditions on individual exemplars within the cluster.

We therefore retain meta-clustering in the main pipeline: although it is not necessary for SEAMS itself, it is necessary for fair comparison against the CS baseline, which depends on stable cluster-level summaries.

\section{Ablation: Selection Mechanism}
\label{sec:ablations}

To support our design choices, we evaluate three SEAMS variants on our proprietary student dataset, each across five seeds. \textbf{Unconstrained retrieval} removes cluster restriction, retrieving over the entire corpus excluding posts by the target author. \textbf{Deterministic selection} removes clipping and noise, selecting the candidate closest to the raw mean embedding. \textbf{Random selection} chooses the final candidate at random.

\begin{table}[t]
\centering
\footnotesize
\setlength{\tabcolsep}{2pt}
\resizebox{\columnwidth}{!}{%
\begin{tabular}{lcccc}
\toprule
\textbf{Method} & \textbf{Sem} & \textbf{Sty} & \textbf{Weg} & \textbf{R-1} \\
\midrule
SEAMS ($M{=}5$) & 87.4 $\pm$ 0.4 & 92.3 $\pm$ 0.4 & 85.3 $\pm$ 1.2 & 64.8 $\pm$ 0.5 \\
Unconstrained & 87.6 $\pm$ 0.3 & 92.5 $\pm$ 0.4 & 85.4 $\pm$ 1.1 & 65.1 $\pm$ 0.4 \\
Deterministic & 87.5 $\pm$ 0.3 & 92.4 $\pm$ 0.4 & 85.3 $\pm$ 1.1 & 64.9 $\pm$ 0.5 \\
Random & 87.2 $\pm$ 0.3 & 91.2 $\pm$ 0.6 & 83.1 $\pm$ 1.5 & 64.2 $\pm$ 0.5 \\
\bottomrule
\end{tabular}}

\vspace{4pt}

\footnotesize
\resizebox{\columnwidth}{!}{%
\begin{tabular}{lcc}
\toprule
\textbf{Method} & \textbf{MATTR} & \textbf{Self-BLEU} \\
\midrule
SEAMS ($M{=}5$) & 85.2 $\pm$ 0.2 & 19.8 $\pm$ 0.2 \\
Unconstrained & 85.2 $\pm$ 0.2 & 19.5 $\pm$ 0.5 \\
Deterministic & 85.2 $\pm$ 0.2 & 19.9 $\pm$ 0.4 \\
Random & 85.2 $\pm$ 0.2 & 20.0 $\pm$ 0.3 \\
\bottomrule
\end{tabular}}
\caption{Generation quality of SEAMS ablation variants on the proprietary dataset, averaged over 5 seeds. All values multiplied by 100.}
\label{tab:ablation-quality}
\end{table}

\begin{table}[t]
\centering
\footnotesize
\setlength{\tabcolsep}{2pt}
\resizebox{\columnwidth}{!}{%
\begin{tabular}{lccccc}
\toprule
\textbf{Method} & \textbf{Weg-ret} & \textbf{Stylo-clf} & \textbf{LUAR} & \textbf{Weg-clf} & \textbf{MPNet-ret} \\
\midrule
SEAMS ($M{=}5$) & 1.89 $\pm$ 0.31 & 2.38 $\pm$ 0.31 & 5.56 $\pm$ 0.73 & 1.35 $\pm$ 0.20 & 5.86 $\pm$ 0.72 \\
Unconstrained & 1.94 $\pm$ 0.26 & 2.49 $\pm$ 0.21 & 5.61 $\pm$ 0.77 & 1.32 $\pm$ 0.25 & 6.05 $\pm$ 0.75 \\
Deterministic & 1.85 $\pm$ 0.31 & 2.37 $\pm$ 0.42 & 5.44 $\pm$ 0.66 & 1.36 $\pm$ 0.24 & 5.87 $\pm$ 0.76 \\
Random & 1.94 $\pm$ 0.26 & 2.32 $\pm$ 0.41 & 5.65 $\pm$ 0.69 & 1.39 $\pm$ 0.36 & 6.02 $\pm$ 0.77 \\
\bottomrule
\end{tabular}}
\caption{Within-cluster authorship attack results for SEAMS ablation variants on the proprietary dataset (MRR ratio, mean $\pm$ std over 5 seeds; lower is harder to attribute; protocol in Section~\ref{sec:authorship}).}
\label{tab:ablation-attack}
\end{table}

All variants except random selection match full SEAMS across the quality distribution (Table~\ref{tab:ablation-quality}), and attack differences remain within seed variance (Table~\ref{tab:ablation-attack}), though unconstrained retrieval is directionally higher on four of five attackers (e.g., $+0.19$ on semantic retrieval).

We draw three findings. First, approximately 71\% of unconstrained top-$K$ retrievals already fall within the target post's cluster, indicating that style retrieval largely recovers cluster membership on its own; cluster restriction provides an in-cluster guarantee that may matter more for smaller or more heterogeneous corpora. Second, random selection degrades quality primarily in the tail: mean style fidelity drops, and more outputs fall below a style fidelity of 80 (12.7\% vs. 10.8\%), with a 10th percentile of 74.4 vs. 78.6. Consensus selection therefore acts as worst-case insurance by filtering degenerate candidates. Third, deterministic selection behaves identically to full SEAMS, but retaining the noise maintains structural consistency with DP-ICL at no measured quality cost.

\section{Embedding-Probe Downstream Evaluation}
\label{app:probe_downstream}

As a complement to fine-tuning (Table~\ref{tab:downstream_utility}), we train logistic regression (ridge regression for Stanford MOOC) on frozen embeddings of the same validation posts, using either a semantic model (\texttt{all-mpnet-base-v2}) or a style model (\texttt{StyleDistance}), and evaluate on the original test posts. For every model, the regularization strength is selected by cross-validation on its training posts alone (5-fold for logistic regression, leave-one-out for ridge regression). Semantic probes lose little with any method's rewrites, while style probes separate the methods: SEAMS or SEAMS-1 has the smallest style loss on every dataset (Table~\ref{tab:downstream_probe}).

\begin{table*}[t]
  \centering
  \small
  \setlength{\tabcolsep}{4pt}
  \begin{tabular}{lcccccc}
    \toprule
    & \multicolumn{2}{c}{\textsc{Proprietary}}
    & \multicolumn{2}{c}{\textsc{Stanford MOOC}}
    & \multicolumn{2}{c}{\textsc{Stack Exchange}} \\
    & \multicolumn{2}{c}{(Bloom's level, macro-F1)}
    & \multicolumn{2}{c}{(confusion, $R^2$)}
    & \multicolumn{2}{c}{(subject, macro-F1)} \\
    \cmidrule(lr){2-3} \cmidrule(lr){4-5} \cmidrule(lr){6-7}
    Method & Semantic $\Delta$ & Style $\Delta$
           & Semantic $\Delta$ & Style $\Delta$
           & Semantic $\Delta$ & Style $\Delta$ \\
    \midrule
    SEAMS     & $-0.6 \pm 2.7$ & $-0.1 \pm 1.0$
              & $+0.5 \pm 2.0$ & $-6.7 \pm 3.7$
              & $-1.1 \pm 0.8$ & $-3.6 \pm 2.7$ \\
    SEAMS-1   & $-0.4 \pm 2.0$ & $\mathbf{+0.0 \pm 1.0}$
              & $+0.6 \pm 1.9$ & $\mathbf{-6.5 \pm 2.0}$
              & $-1.2 \pm 0.7$ & $\mathbf{-3.5 \pm 1.9}$ \\
    ESA       & $+1.4 \pm 2.6$ & $-4.0 \pm 2.6$
              & $+0.9 \pm 1.6$ & $-10.4 \pm 4.7$
              & $-1.0 \pm 0.6$ & $-3.6 \pm 2.8$ \\
    CS        & $+1.3 \pm 1.2$ & $-2.8 \pm 2.8$
              & $+1.0 \pm 1.7$ & $-24.6 \pm 4.9$
              & $-1.0 \pm 1.0$ & $-6.5 \pm 2.5$ \\
    DP-Prompt & $+1.2 \pm 2.1$ & $-16.7 \pm 1.8$
              & $+0.6 \pm 1.2$ & $-16.2 \pm 9.7$
              & $-1.5 \pm 0.9$ & $-7.4 \pm 3.6$ \\
    ZS-CS     & $+0.8 \pm 2.9$ & $-5.6 \pm 1.4$
              & $+0.7 \pm 1.3$ & $-12.0 \pm 5.6$
              & $-1.4 \pm 0.7$ & $-3.6 \pm 2.3$ \\
    ZS-C      & $-0.3 \pm 0.7$ & $-21.1 \pm 4.3$
              & $-0.7 \pm 1.4$ & $-82.9 \pm 58.3$
              & $-1.8 \pm 0.6$ & $-22.3 \pm 6.4$ \\
    \bottomrule
  \end{tabular}
  \caption{%
    Downstream utility with embedding probes (logistic or ridge regression on frozen semantic and style embeddings).
    $\Delta$ = metric when training on rewrites $-$ metric when training on original posts, both tested on original posts; values closer to zero indicate less utility loss.
    Metric is macro-F1 for classification and $R^2$ for regression, $\times$100; mean $\pm$ std over 5 seeds.
    Training on original posts scores 34.7/29.8 (Proprietary), 20.8/17.3 (Stanford MOOC), and 80.5/35.9 (Stack Exchange) with semantic/style embeddings.
    Semantic $\Delta$s are within noise for every method, so bold marks the smallest style loss per dataset only.
  }
  \label{tab:downstream_probe}
\end{table*}

\section{Aggregation Hyperparameters}
\label{app:aggregation}

We use clipping bound $\tau=1$ and noise multiplier $\sigma=0.1$ (Algorithm~\ref{alg:seams}). Candidate embeddings are computed with StyleDistance, whose embeddings have norms of roughly 17, so clipping effectively normalizes each candidate embedding, and the added Gaussian noise has a per-dimension standard deviation of $\sigma\tau/\mathcal{M}=0.02$. The same settings are used for all SEAMS variants that retain noisy aggregation and for the ESA baseline, which shares SEAMS's aggregation step.

\section{Computational Cost}
\label{app:compute}

Clustering is run once per dataset, after an optimal cluster configuration has been selected using a small batch of posts as the validation set. Clustering runs on CPU and takes under 10 minutes per setting for each of the datasets we process.

All experiments using open-source models are conducted on machines with 5 A6000 GPUs (48GB each): 4 host the 72B generation model and 1 hosts the embedding model. We did not record exact wall-clock time, and report file modification times as estimates. On the proprietary dataset, SEAMS ($M=5$) takes approximately 21.6 seconds per post and SEAMS-1 approximately 10.7 seconds per post.

Table~\ref{tab:token-cost} reports token usage per post, computed from the LLM cache produced during our experiments. SEAMS costs $5\times$ more in LLM calls than SEAMS-1, but only about 17\% more in total tokens per post, since each call carries a shorter prompt.

\begin{table}[t]
\centering
\footnotesize
\setlength{\tabcolsep}{3pt}
\resizebox{\columnwidth}{!}{%
\begin{tabular}{lccc}
\toprule
\textbf{Method} & \textbf{Calls} & \textbf{Prompt tok.} & \textbf{Compl. tok.} \\
 & \textbf{/post} & \textbf{/call} & \textbf{/call} \\
\midrule
SEAMS ($M{=}5$) & 5 & $\sim$1,023 & $\sim$105 \\
SEAMS-1 & 1 & $\sim$4,545 & $\sim$259 \\
\bottomrule
\end{tabular}}

\vspace{4pt}

\footnotesize
\resizebox{\columnwidth}{!}{%
\begin{tabular}{lccc}
\toprule
\textbf{Method} & \textbf{Prompt tok.} & \textbf{Compl. tok.} & \textbf{Total tok.} \\
 & \textbf{/post} & \textbf{/post} & \textbf{/post} \\
\midrule
SEAMS ($M{=}5$) & $\sim$5,117 & $\sim$527 & $\sim$5,644 \\
SEAMS-1 & $\sim$4,545 & $\sim$259 & $\sim$4,804 \\
\midrule
Ratio & 1.13$\times$ & 2.03$\times$ & 1.17$\times$ \\
\bottomrule
\end{tabular}}
\caption{Token usage per post on the proprietary dataset. Values are approximate, computed from the LLM cache produced during our experiments.}
\label{tab:token-cost}
\end{table}

\section{SEAMS DSPy Rewriter Specification}
\label{app:dspy_prompt}

The generation module in SEAMS is implemented as a DSPy Signature with three inputs and one output; DSPy compiles it into the model prompt.

\begin{tcblisting}{listing only, breakable, colback=gray!4, colframe=gray!45, boxrule=0.4pt, arc=1mm,
  left=1mm, right=1mm, top=0.5mm, bottom=0.5mm,
  listing options={language=Python, basicstyle=\ttfamily\scriptsize, breaklines=true,
    columns=fullflexible, keepspaces=true, showstringspaces=false}}
class RewriteSignature(dspy.Signature):
    """Generate a student discussion forum post that includes all the key points provided in the `bullet_summary`. Ensure that the post closely mirrors the writing style, tone, and format of the `style_context` exemplar posts. Additionally, follow the structural guidelines detailed in the `post_template` to ensure the rewritten post is well-organized and meets the specified criteria.

    1. **Content Coverage**: Make sure that every point listed in the `bullet_summary` is addressed in the rewritten post.
    2. **Style and Tone**: Adopt the same writing style, tone, and format as the exemplar posts provided in `style_context`. This includes sentence structure, vocabulary, and any specific formatting elements.
    3. **Structure**: Follow the structure outlined in the `post_template`. This may include sections, paragraphs, headings, and other organizational elements that are important for the post's clarity and readability.

    By adhering to these guidelines, the rewritten post should effectively communicate the required content while maintaining a consistent and appropriate style and structure."""

    bullet_summary: str = dspy.InputField(
        desc="Bullet points that the rewritten post must cover.")
    post_template: str = dspy.InputField(
        desc="Description of the general structure of what the rewritten post should look like.")
    style_context: str = dspy.InputField(
        desc="Numbered peer posts demonstrating the target register.")
    rewritten_post: str = dspy.OutputField(
        desc="The rewritten post, content-preserving and register-matched, "
             "with all identifying surface details replaced.")
\end{tcblisting}

\section{Annotation Information}
\label{app:annotation}

\begin{figure*}[t]
    \centering
    \includegraphics[width=\textwidth]{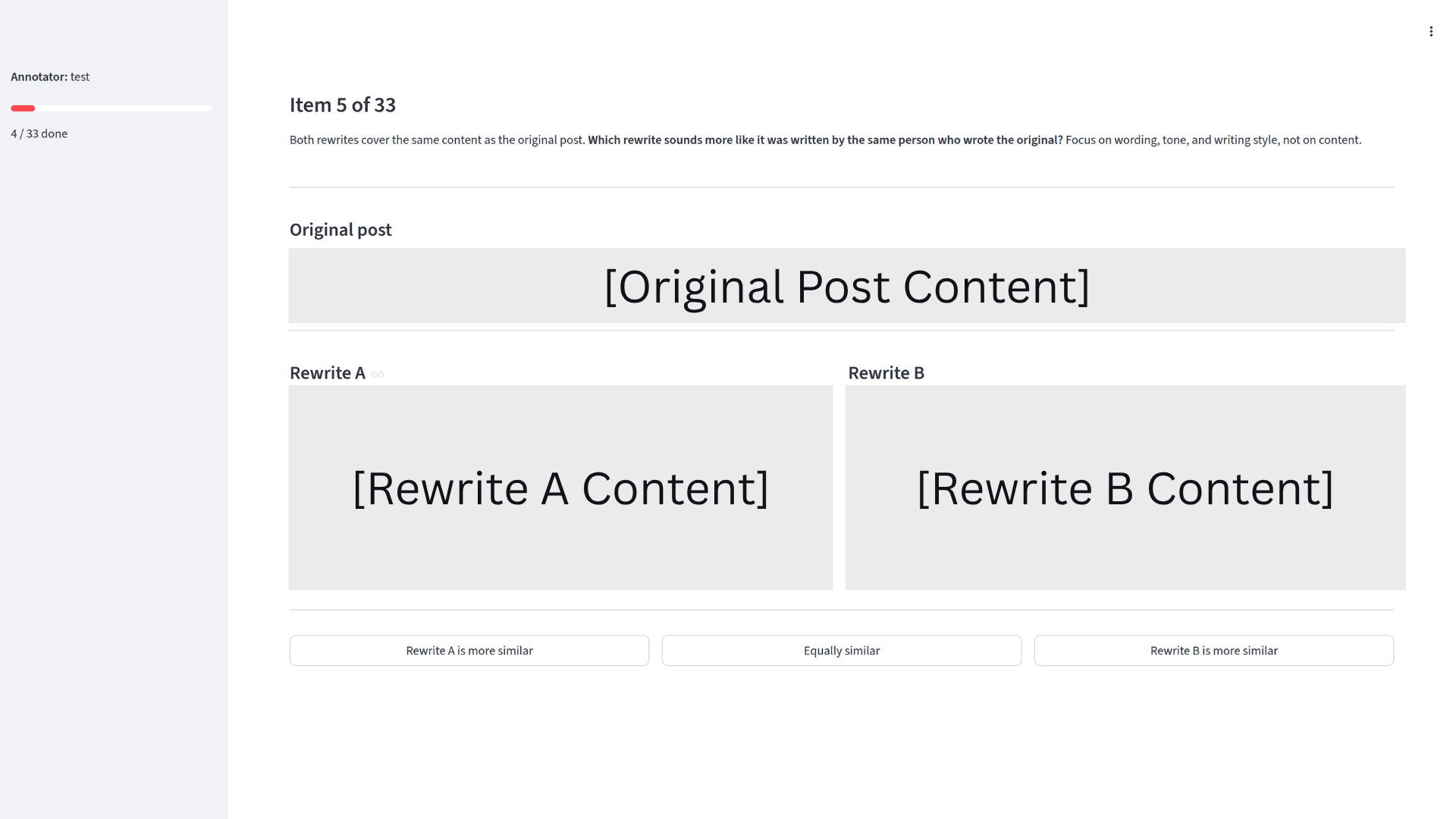}
    \caption{Interface of the style-match evaluation (post text replaced with placeholders).}
    \label{fig:human_eval_annotation}
\end{figure*}

Both human evaluations used a Streamlit application. In the style-match evaluation, each screen showed the original post above two rewrites (``Rewrite A'' and ``Rewrite B'') with the instruction: ``Both rewrites cover the same content as the original post. Which rewrite sounds more like it was written by the same person who wrote the original? Focus on wording, tone, and writing style, not on content.'' Annotators chose ``Rewrite A is more similar,'' ``Equally similar,'' or ``Rewrite B is more similar.'' Three attention checks were placed at fixed positions: a verbatim copy of the original versus scrambled text, an instruction to select a specific option, and a near-copy of a casual post versus unrelated formal text. Figure~\ref{fig:human_eval_annotation} shows the style-match interface.

\section{Naturalness Evaluation}
\label{app:naturalness}

Before the style-match evaluation, we ran a preference-based naturalness study on Prolific with current college and graduate students (\$15/hour). Each pair contained an original post and a generated post with \emph{different} content (from SEAMS, CS, or ZS-CS), along with real-vs.-real control pairs and attention checks, and annotators chose which post ``you or your peers [are] more likely to write,'' or judged them equally natural. After excluding three participants who submitted exceptionally fast or failed attention checks, 26 remained, giving at least five ratings per item. Win rates against original posts (ties = 0.5; $\pm$ half-width of the 95\% item-bootstrap CI) were $0.555 \pm 0.069$ for SEAMS, $0.599 \pm 0.079$ for CS, and $0.547 \pm 0.087$ for ZS-CS, with ordinal Krippendorff's $\alpha = 0.164$. All three methods were judged about as natural as real student posts, and the task did not separate them. Because annotators compare posts with different content and no reference, this design measures general plausibility (and can reward polished text) rather than fidelity to a particular author, which motivated the style-match evaluation in Section~\ref{sec:human-eval}.

\end{document}